\documentclass{article}

\usepackage[final,nonatbib]{nips_2017}

\usepackage[utf8]{inputenc} 
\usepackage[T1]{fontenc}    
\usepackage{hyperref}       
\usepackage{url}            
\usepackage{booktabs}       
\usepackage{amsfonts}       
\usepackage{nicefrac}       
\usepackage{microtype}      

\usepackage{amsmath,amssymb,amsthm}
\usepackage{bm}

\usepackage{times}
\usepackage{graphicx} 
\usepackage{subfigure} 

\usepackage{cite}

\usepackage{algorithm}
\usepackage{algorithmic}
\usepackage{multirow}

\renewcommand{\algorithmicrequire}{\textbf{Input:}}
\renewcommand{\algorithmicensure}{\textbf{Output:}}
\renewcommand\algorithmiccomment[1]{{\small/*{\it#1}*/}}

\theoremstyle{plain}
\newtheorem{thm}{Theorem}[section]

\theoremstyle{definition}
\newtheorem{defn}{Definition}[section]

\theoremstyle{remark}

\title{A Dirichlet Mixture Model of Hawkes Processes for Event Sequence Clustering}

\author{
  Hongteng Xu\thanks{Corresponding author.} \\
  School of ECE\\
  Georgia Institute of Technology\\
  \texttt{hongtengxu313@gmail.com} \\
  \And
  Hongyuan Zha\\
  College of Computing\\
  Georgia Institute of Technology\\
  \texttt{zha@cc.gatech.edu} \\
}

\begin{document}

\maketitle

\begin{abstract}
How to cluster event sequences generated via different point processes is an interesting and important problem in statistical machine learning. 
To solve this problem, we propose and discuss an effective model-based clustering method based on a novel Dirichlet mixture model of a special but significant type of point processes --- Hawkes process. 
The proposed model generates the event sequences with different clusters from the Hawkes processes with different parameters, and uses a Dirichlet distribution as the prior distribution of the clusters. 
We prove the identifiability of our mixture model and propose an effective variational Bayesian inference algorithm to learn our model. 
An adaptive inner iteration allocation strategy is designed to accelerate the convergence of our algorithm.
Moreover, we investigate the sample complexity and the computational complexity  of our learning algorithm in depth. 
Experiments on both synthetic and real-world data show that the clustering method based on our model can learn structural triggering patterns hidden in asynchronous event sequences robustly and achieve superior performance on clustering purity and consistency compared to existing methods.
\end{abstract}

\section{Introduction}\label{sec:intro}
In many practical situations, we need to deal with a huge amount of irregular and asynchronous sequential data.
Typical examples include the viewing records of users in an IPTV system, the electronic health records of patients in hospitals, among many others.  
All of these data are so-called \emph{event sequences}, each of which contains a series of events with different types in the continuous time domain, e.g., when and which TV program a user watched, when and which care unit a patient is transferred to. 
Given a set of event sequences, an important task is learning their clustering structure robustly. 
Event sequence clustering is meaningful for many practical applications. 
Take the previous two examples: clustering IPTV users according to their viewing records is beneficial to the program recommendation system and the ads serving system; 
clustering patients according to their health records helps hospitals to optimize their medication resources.

Event sequence clustering is very challenging. 
Existing work mainly focuses on clustering synchronous (or aggregated) time series with discrete time-lagged observations~\cite{van1999cluster,maharaj2000cluster,liao2005clustering}. 
Event sequences, on the contrary, are in the continuous time domain, so it is difficult to find a universal and tractable representation for them. 
A potential solution is constructing features of event sequences via parametric~\cite{luo2015multi} or nonparametric~\cite{lianmultitask} methods. 
However, these feature-based methods have a high risk of overfitting because of the large number of parameters. 
What is worse, these methods actually decompose the clustering problem into two phases: extracting features and learning clusters. 
As a result, their clustering results are very sensitive to the quality of learned (or predefined) features.

To make concrete progress, we propose a \textbf{D}irichlet \textbf{M}ixture model of \textbf{H}awkes \textbf{P}rocesses (DMHP for short) and study its performance on event sequence clustering in depth. 
In this model, the event sequences belonging to different clusters are modeled via different Hawkes processes. 
The priors of the Hawkes processes' parameters are designed based on their physically-meaningful constraints.
The prior of the clusters is generated via a Dirichlet distribution. 
We propose a variational Bayesian inference algorithm to learn the DMHP model in a nested Expectation-Maximization (EM) framework. 
In particular, we introduce a novel inner iteration allocation strategy into the algorithm with the help of open-loop control theory, which improves the convergence of the algorithm. 
We prove the local identifiability of our model and show that our learning algorithm has better sample complexity and computational complexity than its competitors. 

The contributions of our work include: 
1) We propose a novel Dirichlet mixture model of Hawkes processes and demonstrate its local identifiability. 
To our knowledge, it is the first systematical research on the identifiability problem in the task of event sequence clustering. 
2) We apply an adaptive inner iteration allocation strategy based on open-loop control theory to our learning algorithm and show its superiority to other strategies. 
The proposed strategy achieves a trade-off between convergence performance and computational complexity. 
3) We propose a DMHP-based clustering method. 
It requires few parameters and is robust to the problems of overfitting and model misspecification, which achieves encouraging clustering results.

\section{Related Work}\label{sec:relate}
A temporal point process~\cite{daley2007introduction} is a random process whose realization consists of an event sequence $\{(t_i, c_i)\}_{i=1}^{M}$ with time stamps $t_i \in [0,T]$ and event types $c_i\in \mathcal{C}=\{1,...,C\}$. 
It can be equivalently represented as $C$ counting processes $\{N_c(t)\}_{c=1}^{C}$, where $N_c(t)$ is the number of type-$c$ events occurring at or before time $t$. 
A way to characterize point processes is via the intensity function 
$\lambda_c(t)={\mathbb{E}[dN_c(t)|\mathcal{H}_t^{\mathcal{C}}]}/{dt}$, where $\mathcal{H}_t^{\mathcal{C}}= \{(t_i, c_i) |  t_i < t, c_i\in\mathcal{C}\}$ collects historical events of all types before time $t$. 
It is the expected instantaneous rate of happening type-$c$ events given the history, which captures
the phenomena of interests, i.e., self-triggering~\cite{hawkes1971spectra} or self-correcting~\cite{xu2015trailer}. 

\textbf{Hawkes Processes.} A Hawkes process~\cite{hawkes1971spectra} is a kind of point processes modeling complicated event sequences in which historical events have influences on current and future ones. 
It can also be viewed as a cascade of non-homogeneous Poisson processes~\cite{simma2010modeling,farajtabar2014shaping}. 
We focus on the clustering problem of the event sequences obeying Hawkes processes because Hawkes processes have been proven to be useful for describing real-world data in many applications, e.g., financial analysis~\cite{bacry2012non}, social network analysis~\cite{blundell2012modelling,zhouke2013learning}, system analysis~\cite{luo2015multi}, and e-health~\cite{pmlr-v70-xu17b,reynaud2010adaptive}. 
Hawkes processes have a particular form of intensity:
\begin{eqnarray}\label{dfhp}
\begin{aligned}
\lambda_c(t)=\mu_c + \sideset{}{_{c'=1}^C }\sum\int_{0}^t \phi_{cc'}(s) dN_{c'}(t-s),
\end{aligned}
\end{eqnarray}
where $\mu_c$ is the exogenous base intensity independent of the history while $\sum_{c'=1}^C\int_{0}^t \phi_{cc'}(s) dN_{c'}(t-s)$ the endogenous intensity capturing the peer influence. 
The decay in the influence of historical type-$c'$ events on the subsequent type-$c$ events is captured via the so-called \emph{impact function} $\phi_{cc'}(t)$, which is nonnegative.
A lot of existing work uses predefined impact functions with known parameters, e.g., the exponential functions in~\cite{zhou2013learning,rasmussen2013bayesian} and the power-law functions in~\cite{zhao2015seismic}. 
To enhance the flexibility, a nonparametric model of 1-D Hawkes process was first proposed in~\cite{lewis2011nonparametric} based on ordinary differential equation (ODE) and extended to multi-dimensional case in~\cite{zhouke2013learning,luo2015multi}. 
Another nonparametric model is the contrast function-based model in~\cite{reynaud2010adaptive}, which leads to a Least-Squares (LS) problem~\cite{eichler2016graphical}. 
A Bayesian nonparametric model combining Hawkes processes with infinite relational model is proposed in~\cite{blundell2012modelling}.
Recently, the basis representation of impact functions was used in~\cite{du2012learning,lemonnier2014nonparametric,xu2016learning} to avoid discretization.

\textbf{Sequential Data Clustering and Mixture Models.} Traditional methods mainly focus on clustering synchronous (or aggregated) time series with discrete time-lagged variables~\cite{van1999cluster,maharaj2000cluster,liao2005clustering}. 
These methods rely on probabilistic mixture models~\cite{yakowitz1968identifiability}, extracting features from sequential data and then learning clusters via a Gaussian mixture model (GMM)~\cite{rasmussen1999infinite,maugis2009variable}.
Recently, a mixture model of Markov chains is proposed in~\cite{luo2016learning}, which learns potential clusters from aggregate data. 
For asynchronous event sequences, most of the existing clustering methods can be categorized into feature-based methods, clustering event sequences from learned or predefined features.
Typical examples include the Gaussian process-base multi-task learning method in~\cite{lianmultitask} and the multi-task multi-dimensional Hawkes processes in~\cite{luo2015multi}. 
Focusing on Hawkes processes, the feature-based mixture models in~\cite{li2013dyadic,yang2013mixture,du2015dirichlet} combine Hawkes processes with Dirichlet processes~\cite{blei2006variational,socher2011spectral}. 
However, these methods aim at modeling clusters of events or topics hidden in event sequences (i.e., sub-sequence clustering), which cannot learn clusters of event sequences.
To our knowledge, the model-based clustering method for event sequences has been rarely considered.

\section{Proposed Model}\label{sec:method}
\subsection{Dirichlet Mixture Model of Hawkes Processes}\label{ssec:model}
Given a set of event sequences $\bm{S}=\{\bm{s}_n\}_{n=1}^{N}$, where $\bm{s}_n=\{(t_i, c_i)\}_{i=1}^{M_n}$ contains a series of events $c_i \in\mathcal{C}=\{1,...,C\}$ and their time stamps $t_i \in [0, T_n]$, we model them via a mixture model of Hawkes processes. 
According to the definition of Hawkes process in (\ref{dfhp}), for the event sequence belonging to the $k$-th cluster its intensity function of type-$c$ event at time $t$ is
\begin{eqnarray}\label{lambda}
\begin{aligned}
\lambda_c^k(t)=\mu_c^k + \sideset{}{_{t_i<t}}\sum\phi_{cc_i}^{k}(t-t_i)
		    =\mu_c^k + \sideset{}{_{t_i<t}}\sum\sideset{}{_{d=1}^{D}}\sum a_{cc_id}^{k}g_d(t-t_i),
\end{aligned}
\end{eqnarray}
where $\bm{\mu}^{k}=[\mu_c^k]\in\mathbb{R}_{+}^{C}$ is the exogenous base intensity of the $k$-th Hawkes process.
Following the work in~\cite{xu2016learning}, we represent each impact function $\phi_{cc'}^k(t)$ via basis functions as $\sum_d a_{cc'd}^{k}g_d(t-t_i)$, where $g_d(t)\geq 0$ is the $d$-th basis function and $\bm{A}^k=[a_{cc'd}^{k}]\in\mathbb{R}_{0+}^{C\times C\times D}$ is the coefficient tensor. 
Here we use Gaussian basis function, and their number $D$ can be decided automatically using the basis selection method in~\cite{xu2016learning}. 

In our mixture model, the probability of the appearance of an event sequence $\bm{s}$ is
\begin{eqnarray}\label{mixP}
\begin{aligned}
p(\bm{s};\bm{\Theta}) = \sideset{}{_k}\sum\pi^k \mbox{HP}(\bm{s}|\bm{\mu}^k, \bm{A}^k),~
\mbox{HP}(\bm{s}|\bm{\mu}^k, \bm{A}^k)
=\sideset{}{_i}\prod \lambda_{c_i}^{k}(t_i)\exp\Bigl(-\sideset{}{_c}\sum\int_0^T\lambda_c^k(s)ds\Bigr).
\end{aligned}
\end{eqnarray}
Here $\pi^k$'s are the probabilities of clusters and $\mbox{HP}(\bm{s}|\bm{\mu}^k, \bm{A}^k)$ is the conditional probability of the event sequence $\bm{s}$ given the $k$-th Hawkes process, which follows the intensity function-based definition in~\cite{daley2007introduction}. 
According to the Bayesian graphical model, we regard the parameters of Hawkes processes, $\{\bm{\mu}^k, \bm{A}^k\}$, as random variables. 
For $\bm{\mu}^k$'s, we consider its positiveness and assume that they obey $C\times K$ independent Rayleigh distributions. 
For $\bm{A}^{k}$'s, we consider its nonnegativeness and sparsity as the work in~\cite{zhou2013learning,luo2015multi,xu2016learning}) did, and assume that they obey $C\times C\times D\times K$ independent exponential distributions. 
The prior of cluster is a Dirichlet distribution. 
Therefore, we can describe the proposed Dirichlet mixture model of Hawkes process in a generative way as
\begin{eqnarray*}
\begin{aligned}
&\bm{\pi}\sim\mbox{Dir}(\alpha / K,...,\alpha / K),~k|\bm{\pi}\sim\mbox{Category}(\bm{\pi}),\\
&\bm{\mu}\sim\mbox{Rayleigh}(\bm{B}),~\bm{A}\sim\mbox{Exp}(\bm{\Sigma}),~\bm{s}|k,\bm{\mu},\bm{A}\sim\mbox{HP}(\bm{\mu}_k,\bm{A}_k),
\end{aligned}
\end{eqnarray*} 
Here $\bm{\mu}=[\mu_{c}^{k}]\in\mathbb{R}_{+}^{C\times K}$ and $\bm{A}=[a_{cc'd}^{k}]\in\mathbb{R}_{0+}^{C\times C\times D\times K}$ are parameters of Hawkes processes, and 
$\{\bm{B}=[\beta_{c}^{k}], \bm{\Sigma}=[\sigma_{cc'd}^{k}]\}$ are hyper-parameters. 
Denote the latent variables indicating the labels of clusters as matrix $\bm{Z}\in \{0,1\}^{N\times K}$. 
We can factorize the joint distribution of all variables as\footnote{$\mbox{Rayleigh}(x|\beta)=\frac{x}{\beta^2}e^{-\frac{x^2}{2\beta^2}}$, $\mbox{Exp}(x|\sigma)=\frac{1}{\sigma}e^{-\frac{x}{\sigma}}$, $x\geq 0$.} 
\begin{eqnarray}\label{factorize}
\begin{aligned}
&p(\bm{S},\bm{Z},\bm{\pi},\bm{\mu},\bm{A})=p(\bm{S}|\bm{Z},\bm{\mu},\bm{A})p(\bm{Z}|\bm{\pi})p(\bm{\pi})p(\bm{\mu})p(\bm{A}),~\text{where}\\
&p(\bm{S}|\bm{Z},\bm{\mu},\bm{A})=\sideset{}{_{n,k}}\prod \mbox{HP}(\bm{s}_n|\bm{\mu}^k, \bm{A}^k)^{z_{nk}},\quad p(\bm{Z}|\bm{\pi})=\sideset{}{_{n,k}}\prod(\pi^k)^{z_{nk}},\\
&p(\bm{\pi})=\mbox{Dir}(\bm{\pi}|\bm{\alpha}),\quad p(\bm{\mu})=\sideset{}{_{c,k}}\prod\mbox{Rayleigh}(\mu_c^k|\beta_{c}^k),\quad p(\bm{A})=\sideset{}{_{c,c',d,k}}\prod\mbox{Exp}(a_{cc'd}^k|\sigma_{cc'd}^k).
\end{aligned}
\end{eqnarray}

Our mixture model of Hawkes processes are different from the models in~\cite{li2013dyadic,yang2013mixture,du2015dirichlet}. 
Those models focus on the sub-sequence clustering problem within an event sequence. 
The intensity function is a weighted sum of multiple intensity functions of different Hawkes processes. 
Our model, however, aims at finding the clustering structure across different sequences. 
The intensity of each event is generated via a single Hawkes process, while the likelihood of an event sequence is a mixture of likelihood functions from different Hawkes processes. 

\subsection{Local Identifiability} 
One of the most important questions about our mixture model is whether it is identifiable or not. 
According to the definition of Hawkes process and the work in~\cite{rothenberg1971identification,meijer2008simple}, we can prove that our model is locally identifiable. The proof of the following theorem is given in the supplementary file. 
\begin{thm}\label{the1}
When the time of observation goes to infinity, the mixture model of the Hawkes processes defined in~(\ref{mixP}) is locally identifiable, i.e., for each parameter point $\bm{\Theta}=\mbox{vec}\left(\tiny{\begin{bmatrix}
\pi^1&...&\pi^K\\ \bm{\theta}^1&...&\bm{\theta}^K
\end{bmatrix}}\right)$, where $\bm{\theta}^k=\{\bm{\mu}^k,\bm{A}^k\}\in \mathbb{R}_{+}^C\times\mathbb{R}_{0+}^{C\times C\times D}$ for $k=1,..,K$, there exists an open neighborhood of $\bm{\Theta}$ containing no other $\bm{\Theta}'$ which makes $p(\bm{s};\bm{\Theta})=p(\bm{s};\bm{\Theta}')$ holds for all possible $\bm{s}$.
\end{thm}

\section{Proposed Learning Algorithm}
\subsection{Variational Bayesian Inference}\label{ssec:alg}
Instead of using purely MCMC-based learning method like~\cite{rasmussen2013bayesian}, we propose an effective variational Bayesian inference algorithm to learn (\ref{factorize}) in a nested EM framework.
Specifically, we consider a variational distribution having the following factorization:
\begin{eqnarray}\label{surrogate}
\begin{aligned}
q(\bm{Z},\bm{\pi},\bm{\mu},\bm{A})=q(\bm{Z})q(\bm{\pi},\bm{\mu},\bm{A})=q(\bm{Z})q(\bm{\pi})\sideset{}{_k}\prod q(\bm{\mu}^k)q(\bm{A}^k).
\end{aligned}
\end{eqnarray}
An EM algorithm can be used to optimize (\ref{surrogate}). 

\textbf{Update Responsibility (E-step).} The logarithm of the optimized factor $q^{*}(\bm{Z})$ is approximated as
\begin{eqnarray*}\label{qZ}
\begin{aligned}
&\log q^*(\bm{Z})
=\mathbb{E}_{\bm{\pi}}[\log p(\bm{Z}|\bm{\pi})]+\mathbb{E}_{\bm{\mu},\bm{A}}[\log p(\bm{S}|\bm{Z},\bm{\mu},\bm{A})]+\mathsf{C}\\
=&\sideset{}{_{n,k}}\sum z_{nk}\left(\mathbb{E}[\log \pi^k]+\mathbb{E}[\log\mbox{HP}(\bm{s}_n|\bm{\mu}^k,\bm{A}^k)]\right)+\mathsf{C}\\
=&\sideset{}{_{n,k}}\sum z_{nk}\Bigl(\mathbb{E}[\log \pi^k]+\mathbb{E}[\sideset{}{_i}\sum\log\lambda_{c_i}^{k}(t_i)-\sideset{}{_c}\sum\int_0^{T_n}\lambda_c^k(s)ds]\Bigr)+\mathsf{C}\\
\approx &\sideset{}{_{n,k}}\sum z_{nk}\underbrace{\Bigl(\mathbb{E}[\log \pi^k]+\sideset{}{_i}\sum\Bigl(\log\mathbb{E}[\lambda_{c_i}^{k}(t_i)]-\frac{\text{Var}[\lambda_{c_i}^{k}(t_i)]}{2\mathbb{E}^2[\lambda_{c_i}^{k}(t_i)]}\Bigr)-\sideset{}{_c}\sum\mathbb{E}[\int_0^{T_n}\lambda_c^k(s)ds]\Bigr)}_{\rho_{nk}}+\mathsf{C}.
\end{aligned}
\end{eqnarray*}
where $\mathsf{C}$ is a constant and $\mbox{Var}[\cdot]$ represents the variance of random variable. 
Each term $\mathbb{E}[\log\lambda_{c}^{k}(t)]$ is approximated via its second-order Taylor expansion $\log\mathbb{E}[\lambda_{c}^{k}(t)]-\frac{\text{Var}[\lambda_{c}^{k}(t)]}{2\mathbb{E}^2[\lambda_{c}^{k}(t)]}$~\cite{teh2006collapsed}.
Then, the responsibility $r_{nk}$ is calculated as
\begin{eqnarray}\label{response}
\begin{aligned}
r_{nk}=\mathbb{E}[z_{nk}]={\rho_{nk}}/{(\sideset{}{_j}\sum\rho_{nj})}.
\end{aligned}
\end{eqnarray}
Denote $N_k=\sum_{n}r_{nk}$ for all $k$'s. 

\textbf{Update Parameters (M-step).} The logarithm of optimal factor $q^*(\bm{\pi},\bm{\mu},\bm{A})$ is 
\begin{eqnarray*}\label{qAmu}
\begin{aligned}
&\log q^*(\bm{\pi},\bm{\mu},\bm{A})\\
=&\sideset{}{_k}\sum\log(p(\bm{\mu}^k)p(\bm{A}^k)) + \mathbb{E}_{\bm{Z}}[\log p(\bm{Z}|\bm{\pi})]
+\log p(\bm{\pi})+\sideset{}{_{n,k}}\sum r_{nk}\log \mbox{HP}(\bm{s}_n|\bm{\mu}^k,\bm{A}^k)+\mathsf{C}.
\end{aligned}
\end{eqnarray*}
We can estimate the parameters of Hawkes processes via:
\begin{eqnarray}\label{MLE}
\begin{aligned}
\hat{\bm{\mu}},\widehat{\bm{A}}=\arg\sideset{}{_{\bm{\mu},\bm{A}}}\max~\log(p(\bm{\mu})p(\bm{A}))+\sideset{}{_{n,k}}\sum r_{nk}\log \mbox{HP}(\bm{s}_n|\bm{\mu}^k,\bm{A}^k).
\end{aligned}
\end{eqnarray}
Following the work in~\cite{zhou2013learning,yang2013mixture,xu2016learning}, 
we need to apply an EM algorithm to solve (\ref{MLE}) iteratively. 
After getting optimal $\hat{\bm{\mu}}$ and $\widehat{\bm{A}}$, we update distributions as
\begin{eqnarray}\label{update}
\begin{aligned}
\bm{\Sigma}^k = \widehat{\bm{A}}^k,~
\bm{B}^k =\sqrt{2/\pi}\hat{\bm{\mu}}^k.
\end{aligned}
\end{eqnarray}

\textbf{Update The Number of Clusters $K$.} 
When the number of clusters $K$ is unknown, we initialize $K$ randomly and update it in the learning phase. 
There are multiple methods to update the number of clusters. 
Regrading our Dirichlet distribution as a finite approximation of a Dirichlet process, we set a large initial $K$ as the truncation level.
A simple empirical method is discarding the empty cluster (i.e., $N_k=0$) and merging the cluster with $N_k$ smaller than a threshold $N_{min}$ in the learning phase. 
Besides this, we can apply the MCMC in~\cite{green1995reversible,zhang2004learning} to update $K$ via merging or splitting clusters. 

Repeating the three steps above, our algorithm maximizes the log-likelihood function (i.e., the logarithm of (\ref{factorize})) and achieves optimal $\{\bm{\Sigma}, \bm{B}\}$ accordingly. 
Both the details of our algorithm and its computational complexity are given in the supplementary file. 

\subsection{Inner Iteration Allocation Strategy and Convergence Analysis}
Our algorithm is in a nested EM framework, where the outer iteration corresponds to the loop of E-step and M-step and the inner iteration corresponds to the inner EM in the M-step. 
The runtime of our algorithm is linearly proportional to the total number of inner iterations. 
Given fixed runtime (or the total number of inner iterations), both the final achievable log-likelihood and convergence behavior of the algorithm highly depend on how we allocate the number of inner iterations across the outer iterations. 
In this work, we test three inner iteration allocation strategies. 
The first strategy is \emph{heuristic}, which fixes, increases, or decreases the number of inner iterations as the outer iteration progresses. 
Compared with constant inner iteration strategy, the increasing or decreasing strategy might improve the convergence of algorithm~\cite{golub2000large}. 
The second strategy is based on \emph{open-loop control}~\cite{ogunnaike1994process}: in each outer iteration, we compute objective function via two methods respectively --- updating parameters directly (i.e., continuing current M-step and going to next inner iteration) or first updating responsibilities and then updating parameters (i.e., going to a new loop of E-step and M-step and starting a new outer iteration). 
The parameters corresponding to the smaller negative log-likelihood are preserved. 
The third strategy is applying \emph{Bayesian optimization}~\cite{snoek2012practical,shahriari2016taking} to optimize the number of inner iterations per outer iteration via maximizing the expected improvement. 

\begin{figure}[t]
\centering
\subfigure[Random Sparse Coefficients]{
\includegraphics[width=1\linewidth]{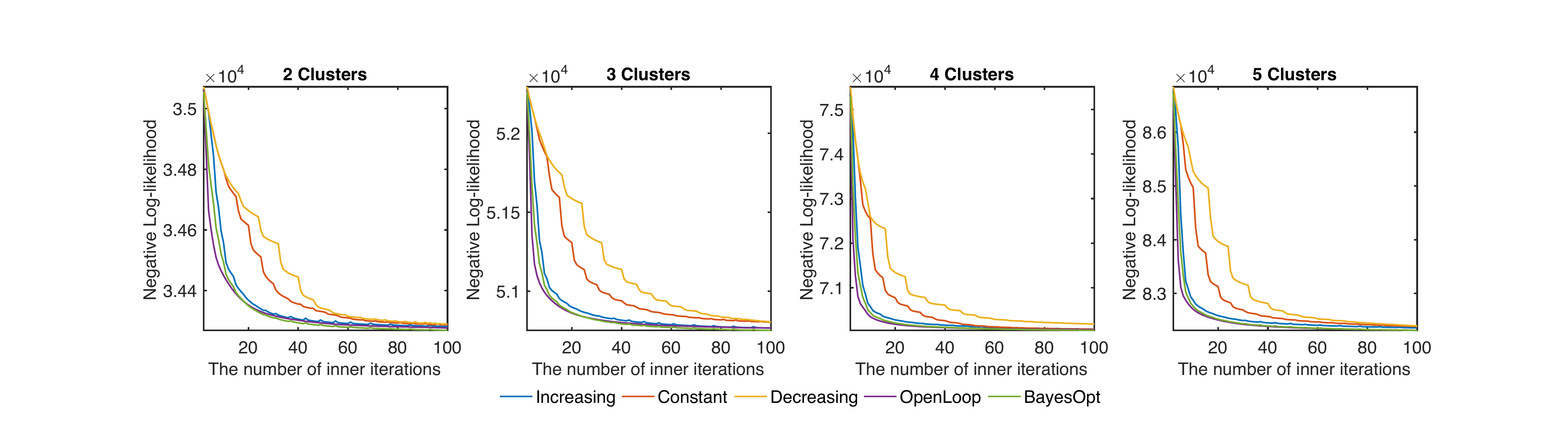}\label{fig:c1}
}
\subfigure[Blockwise Sparse Coefficients]{
\includegraphics[width=1\linewidth]{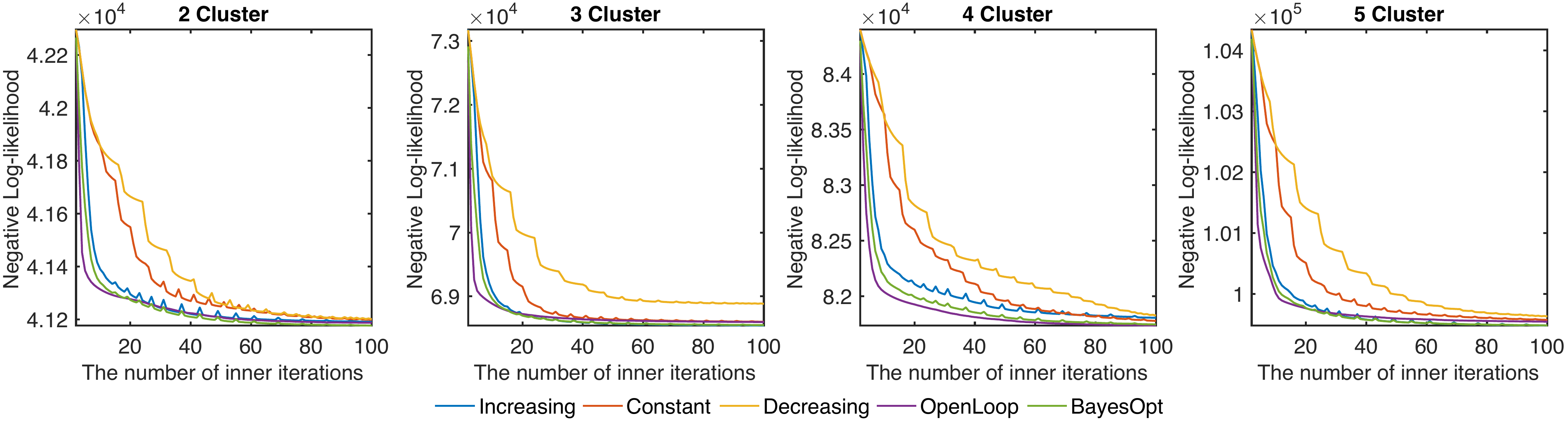}\label{fig:c2}
}\vspace{-10pt}
\caption{Comparison for various inner iteration allocation strategies on different synthetic data sets. 
Each curve is the average of $5$ trials' results. 
In each trial, total $100$ inner iterations are applied. 
The increasing (decreasing) strategy changes the number of inner iterations from $2$ to $8$ (from $8$ to $2$). 
The constant strategy fixes the number to $5$. }\label{fig:conv1}
\end{figure}

\begin{figure}[t!]
\centering
\includegraphics[width=0.91\linewidth]{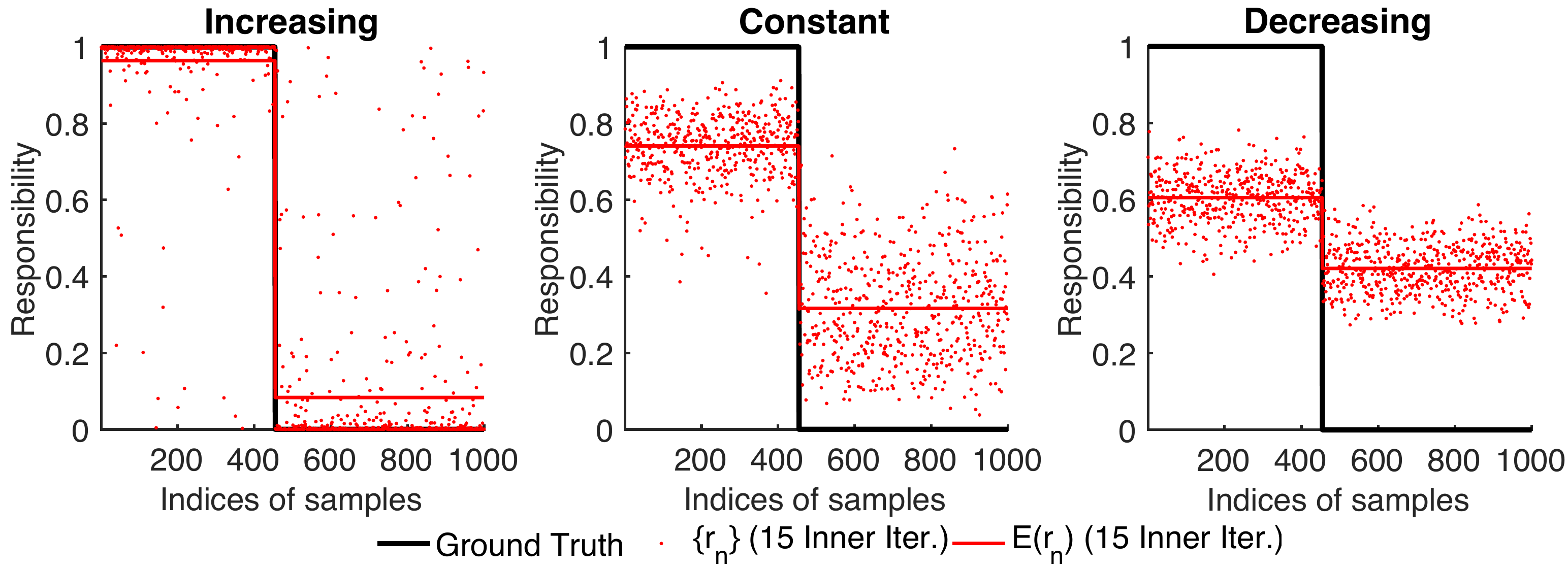}
\vspace{-10pt}
\caption{The data contain $200$ event sequences generated via two $5$-dimensional Hawkes processes.  
The black line is the ground truth. 
The red dots are responsibilities after $15$ inner iterations, and the red line is their average.}\label{fig:conv2}
\vspace{-3pt}
\end{figure}

We apply these strategies to $8$ synthetic data sets and visualize their impacts on the convergence of our algorithm in Fig.~\ref{fig:conv1}. 
All the data sets are generated by the Hawkes processes with sparse coefficients $\bm{A}$. 
In the first four data sets, the nonzero elements in $\bm{A}$ are distributed randomly, and the number of clusters increases from $2$ to $5$. 
In the last four data sets, each slide of $\bm{A}^k$, $k=1,...,K$, contain several all-zero columns and rows (i.e. blockwise sparse tensor). 
In Fig.~\ref{fig:conv1}, we can find that the open-loop control strategy and the Bayesian optimization strategy obtain comparable performance on the convergence of algorithm. 
Both of them outperform heuristic strategies (i.e., increasing, decreasing and fixing the number of inner iterations per outer iteration), which reduce the negative log-likelihood more rapidly and reach lower value finally. 
Although adjusting the number of inner iterations via different methodologies, both these two strategies tend to increase the number of inner iterations w.r.t. the number of outer iterations. 
In the beginning of algorithm, the open-loop control strategy updates responsibilities frequently, and similarly, the Bayesian optimization strategy assigns small number of inner iterations. 
The heuristic strategy that increasing the number of inner iterations follows the same tendency, and therefore, is just slightly worse than the open-loop control and the Bayesian optimization. 
This phenomenon is because the estimated responsibility is not reliable in the beginning. 
Too many inner iterations at that time might make learning results fall into bad local optimums. 

Fig.~\ref{fig:conv2} further verifies our explanation. 
With the help of the increasing strategy, most of the responsibilities converge to the ground truth with high confidence after just $15$ inner iterations, because the responsibilities has been updated over $5$ times.
On the contrary, the responsibilities corresponding to the constant and the decreasing strategies have more uncertainty --- many responsibilities are around $0.5$ and far from the ground truth. 

Based on the analysis above, the {\it increasing} allocation strategy indeed improves the convergence of our algorithm, and the open-loop control and the Bayesian optimization are superior to other competitors. 
Because the computational complexity of the open-loop control is much lower than that of the Bayesian optimization, in the following experiments, we apply open-loop control strategy to our learning algorithm. 

\begin{algorithm}[t]
   \caption{Learning DMHP}
   \label{alg1}
\begin{algorithmic}[1]
   \STATE \textbf{Input:} $\bm{S}=\{\bm{s}_n\}_{n=1}^{N}$, the maximum number of clusters $K$, the maximum number of iteration $I$.
   \STATE \textbf{Output:} Optimal parameters of model, $\hat{\bm{\alpha}}$, $\widehat{\bm{\Sigma}}$, and $\widehat{\bm{B}}$.
   \STATE Initialize $\bm{\alpha}$, $\bm{\Sigma}$, $\bm{B}$ and $[r_{nk}]$ randomly, $i=0$.
   \REPEAT
   \STATE \textbf{Just \emph{M-step}:}
   \STATE Given $[r_{nk}]$, update $\{\hat{\bm{\mu}}^{(1)},\widehat{\bm{A}}^{(1)}\}$ via~(\ref{Amu}), calculate negative log-likelihood $L^{(1)}$.
   \STATE \textbf{A loop of \emph{E-step and M-step}:}
   \STATE Given $\{\bm{\alpha}, \bm{\Sigma}, \bm{B}\}$, update responsibility via~(\ref{response}), denoted as $[r_{nk}^2]$ . 
   \STATE Given $[r_{nk}^2]$, update $\{\hat{\bm{\mu}}^{(2)},\widehat{\bm{A}}^{(2)}\}$ via~(\ref{Amu}), calculate negative log-likelihood $L^{(2)}$.
   \STATE \textbf{If} $L^{(1)}<L^{(2)}$
   \STATE \quad Given $\{\hat{\bm{\mu}}^{(1)},\widehat{\bm{A}}^{(1)}\}$, update $\bm{\Sigma}$, $\bm{B}$ via~(\ref{update}).
   \STATE \textbf{Else}
   \STATE \quad Update $[r_{nk}]$ via $[r_{nk}^{(2)}]$.
   \STATE \quad Given $[r_{nk}], \hat{\bm{\mu}}^{(2)},\widehat{\bm{A}}^{(2)}$, update $\bm{\alpha}$, $\bm{\Sigma}$, $\bm{B}$ via~(\ref{update}).   
   \STATE \textbf{End} 
   \STATE Merge or split clusters and update $\bm{\Sigma}$, $\bm{B}$ via MCMC.
   \STATE $i=i+1$.
   \UNTIL{$i=I$} 
   \STATE $\hat{\bm{\alpha}}=\bm{\alpha}$, $\widehat{\bm{\Sigma}}=\bm{\Sigma}$, and $\widehat{\bm{B}}=\bm{B}$. 
\end{algorithmic}
\end{algorithm}

\subsection{Computational Complexity and Acceleration}
Given $N$ training sequences of $C$-dimensional Hawkes processes, each of which contains $I$ events, we represent impact functions by $D$ basis functions and set the maximum number of clusters to be $K$. 
In the worst case, the computational complexity per iteration of our learning algorithm is $\mathcal{O}(KDNI^3C^2)$. 
Fortunately, the exponential prior of tensor $\bm{A}$ corresponds to a sparse regularizer. 
In the learning phase, we can ignore the computations involving the elements close to zero to reduce the computational complexity. 
If the number of nonzero elements in each $\bm{A}^{k}$ is comparable to $C$, then the computational complexity of our algorithm will be $\mathcal{O}(KDNI^2C)$. 
Additionally, the parallel computing techniques can also be applied to further reduce the runtime of our algorithm. 
Note that the learning algorithm of MMHP discretizes each impact function into $L$ points and estimates them via finite element analysis. 
The low-rank regularizer is imposed on its parameters.
Therefore, its computational complexity per iteration is $\mathcal{O}(NI(I^2C^2+L(C+I))+C^3)$. 
Similarly, when the parameters of each Hawkes process is sparse, its computational complexity will reduce to $\mathcal{O}(NI(IC+L(C+I))+C^2)$. 
The first part $\mathcal{O}(NI(IC+L(C+I)))$ corresponds to the ODE-based parameter updating while the second part $\mathcal{O}(C^2)$ corresponds to the soft-thresholding of parameters. 
According to the setting in~\cite{zhouke2013learning,luo2015multi}, generally $L\gg I$.
Therefore, the computational complexity of our algorithm is superior to that of MMHP, especially in high dimensional cases (i.e., large $C$).

\subsection{Empirical Analysis of Sample Complexity}
Focusing on the task of clustering event sequences, we investigate the sample complexity of our DMHP model and its learning algorithm. 
In particular, we want to show that the clustering method based on our model requires fewer samples than existing methods to identify clusters successfully. 
Among existing methods, the main competitor of our method is the clustering method based on the multi-task multi-dimensional Hawkes process (MMHP) model in~\cite{luo2015multi}.
It learns a specific Hawkes process for each sequence and clusters the sequences via applying the Dirichlet processes Gaussian mixture model (DPGMM)~\cite{rasmussen1999infinite,gorur2010dirichlet} to the parameters of the corresponding Hawkes processes.

\begin{figure}[t]
\centering
\subfigure[MMHP+DPGMM]{
\includegraphics[height=4cm]{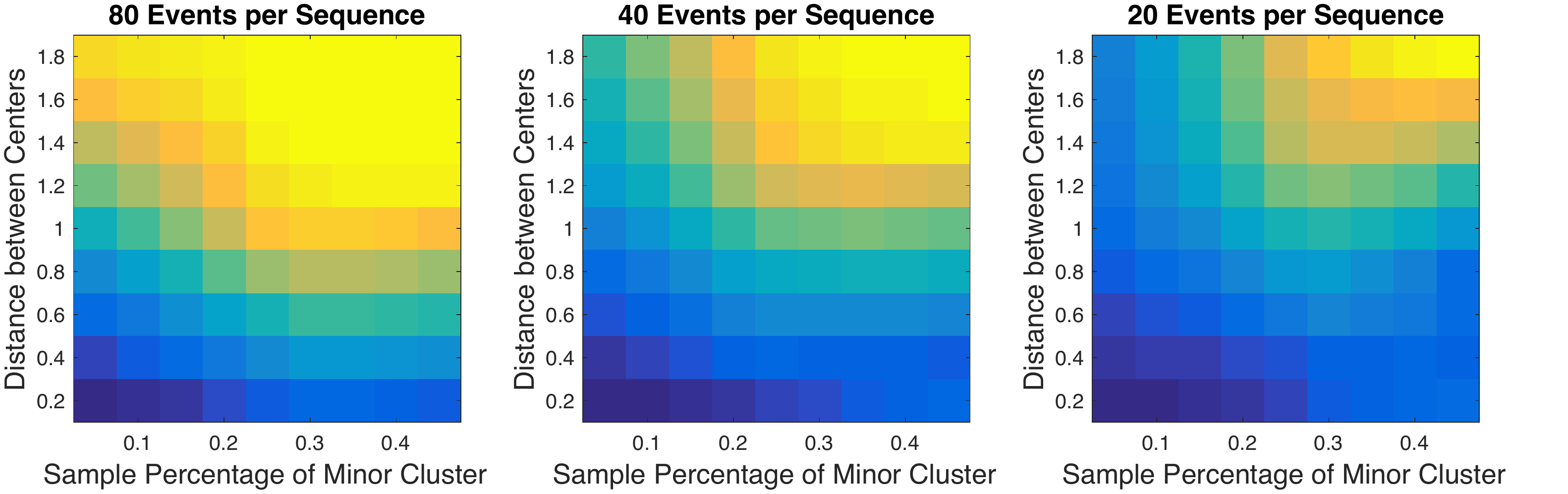}\label{fig:EOI2}
}
\subfigure[DMHP]{
\includegraphics[height=4cm]{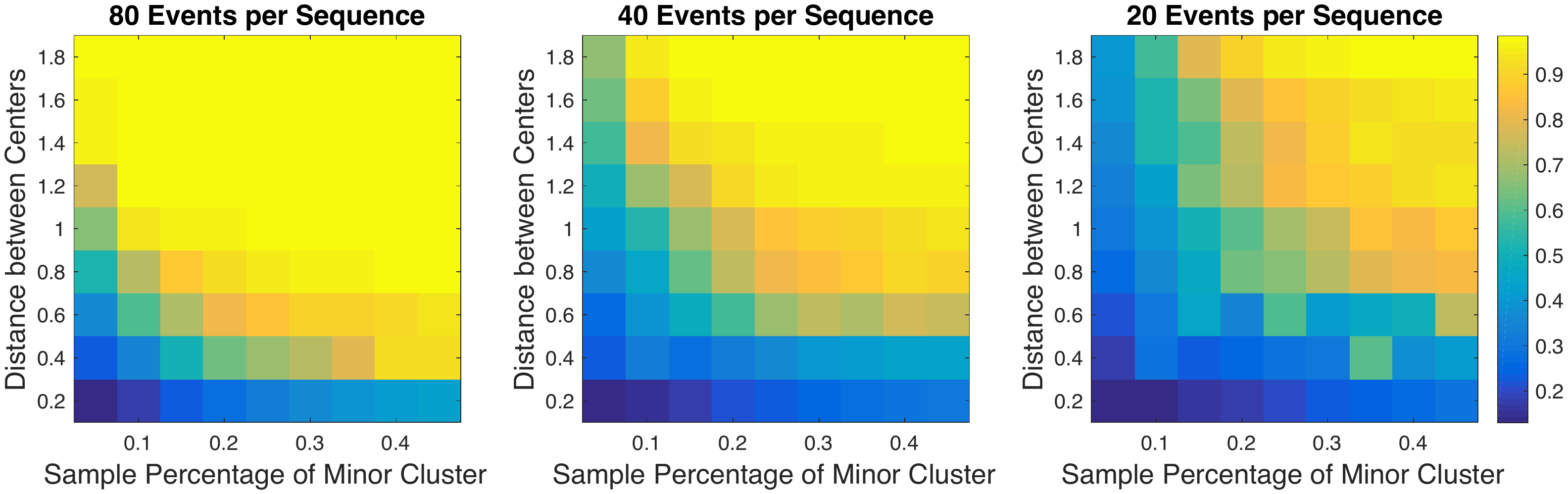}\label{fig:EOI1}
}
\vspace{-10pt}
\caption{Comparisons for various methods on F1 score of minor cluster. }\label{fig:eoi}
\vspace{-3pt}
\end{figure}

Following the work in~\cite{kim2008mixture}, we demonstrate the superiority of our DMHP-based clustering method through the comparison on the identifiability of minor clusters given finite number of samples. 
Specifically, we consider a binary clustering problem with $500$ event sequences. 
For the $k$-th cluster, $k=1,2$, $N_k$ event sequences are generated via a $1$-dimensional Hawkes processes with parameter $\bm{\theta}^k=\{\bm{\mu}^k,\bm{A}^k\}$. 
Taking the parameter as a representation of the clustering center, we can calculate the distance between two clusters as $d=\|\bm{\theta}^1-\bm{\theta}^2\|_2$. 
Assume that $N_1<N_2$, we denote the first cluster as ``minor'' cluster, whose sample percentage is $\pi^1=\frac{N_1}{N_1+N_2}$.  
Applying our DMHP model and its learning algorithm to the data generated with different $d$'s and $\pi^1$'s, we can calculate the F1 scores of the minor cluster w.r.t. $\{d, \pi\}$. 
The high F1 score means that the minor cluster is identified with high accuracy. 
Fig.~\ref{fig:eoi} visualizes the maps of F1 scores generated via different methods w.r.t. the number of events per sequence.    
We can find that the F1 score obtained via our DMHP-based method is close to $1$ in most situations. 
Its identifiable area (yellow part) is much larger than that of the MMHP+DPGMM method consistently w.r.t. the number of events per sequence. 
The unidentifiable cases happen only in the following two situations: the parameters of different clusters are nearly equal (i.e., $d\rightarrow 0$); or the minor cluster is extremely small (i.e., $\pi^1\rightarrow 0$). 
The enlarged version of Fig.~\ref{fig:eoi} is given in the supplementary file.

\section{Experiments}\label{sec:exp}
To demonstrate the feasibility and the efficiency of our \textbf{DMHP}-based sequence clustering method, we compare it with the state-of-the-art methods, including the vector auto-regressive (\textbf{VAR}) method~\cite{han2013transition}, the Least-Squares (\textbf{LS}) method in~\cite{eichler2016graphical}, and the multi-task multi-dimensional Hawkes process (\textbf{MMHP}) in~\cite{luo2015multi}.  
All of the three competitors first learn features of sequences and then apply the \textbf{DPGMM}~\cite{gorur2010dirichlet} to cluster sequences. 
The VAR discretizes asynchronous event sequences to time series and learns transition matrices as features. 
Both the LS and the MMHP learn a specific Hawkes process for each event sequence. 
For each event sequence, we calculate its infectivity matrix $\bm{\Phi}=[\phi_{cc'}]$, where the element $\phi_{cc'}$ is the integration of impact function (i.e., $\int_{0}^{\infty}\phi_{cc'}(t)dt$), and use it as the feature. 

For the synthetic data with clustering labels, we use \emph{clustering purity}~\cite{manning2008introduction} to evaluate various methods:
\begin{eqnarray*}
\begin{aligned}
\mbox{Purity}=\frac{1}{N}\sideset{}{_{k=1}^{K}}\sum\sideset{}{_{j\in\{1,...,K'\}}}\max|\mathcal{W}_k\cap \mathcal{C}_j|,
\end{aligned}
\end{eqnarray*}
where $\mathcal{W}_k$ is the learned index set of sequences belonging to the $k$-th cluster, $\mathcal{C}_j$ is the real index set of sequence belonging to the $j$-th class, and $N$ is the total number of sequences. 
For the real-world data, we visualize the infectivity matrix of each cluster and measure the \emph{clustering consistency} via a cross-validation method~\cite{tibshirani2005cluster,von2010clustering}. 
The principle is simple: because random sampling does not change the clustering structure of data, a clustering method with high consistency should preserve the pairwise relationships of samples in different trials.
Specifically, we test each clustering method with $J$ ($=100$) trials. 
In the $j$-th trial, data is randomly divided into two folds. 
After learning the corresponding model from the training fold, we apply the method to the testing fold. 
We enumerate all pairs of sequences within a same cluster in the $j$-th trial and count the pairs preserved in all other trials. 
The clustering consistency is the minimum proportion of preserved pairs over all trials:
\begin{eqnarray*}
\begin{aligned}
\mbox{Consistency}=\sideset{}{_{j\in\{1,..,J\}}}\min\sideset{}{_{j'\neq j}}\sum\sideset{}{_{(n,n')\in \mathcal{M}_j}}\sum\frac{1\{k_n^{j'}=k_{n'}^{j'}\}}{(J-1)|\mathcal{M}_j|},
\end{aligned}
\end{eqnarray*}
where $\mathcal{M}_j=\{(n,n')|k_n^j=k_{n'}^j\}$ is the set of sequence pairs within same cluster in the $j$-th trial, and $k_n^j$ is the index of cluster of the $n$-th sequence in the $j$-th trial.

\begin{table}[t]
  \centering
  \caption{Clustering Purity on Synthetic Data.\label{tab1}}\vspace{-5pt}
    \begin{small}
      \begin{tabular}{
        @{\hspace{5pt}}c@{\hspace{5pt}}|
        @{\hspace{5pt}}c@{\hspace{5pt}}|
        @{\hspace{5pt}}c@{\hspace{5pt}}
        @{\hspace{5pt}}c@{\hspace{5pt}}
        @{\hspace{5pt}}c@{\hspace{5pt}}
        @{\hspace{5pt}}c@{\hspace{5pt}}
        @{\hspace{5pt}}|c@{\hspace{5pt}}
        @{\hspace{5pt}}c@{\hspace{5pt}}
        @{\hspace{5pt}}c@{\hspace{5pt}}
        @{\hspace{5pt}}c@{\hspace{5pt}}
        }
        \hline\hline
        &&\multicolumn{4}{c|}{Sine-like $\phi(t)$}&\multicolumn{4}{c}{Piecewise constant $\phi(t)$}\\ 
        \hline
        \multirow{2}{*}{$C$} &\multirow{2}{*}{$K$} &VAR+ &LS+ &MMHP+ &\multirow{2}{*}{DMHP} &VAR+ &LS+ &MMHP+ &\multirow{2}{*}{DMHP}\\
        & &DPGMM &DPGMM &DPGMM & &DPGMM &DPGMM &DPGMM \\
        \hline
        \multirow{4}{*}{5} &2 &0.5235 &0.5639 &0.5917 &\textbf{0.9898}
                              &0.5222 &0.5589 &0.5913 &\textbf{0.8085}\\
                           &3 &0.3860 &0.5278 &0.5565 &\textbf{0.9683}
                              &0.3618 &0.4402 &0.4517 &\textbf{0.7715}\\
                           &4 &0.2894 &0.4365 &0.5112 &\textbf{0.9360}
                              &0.2901 &0.3365 &0.3876 &\textbf{0.7056}\\
                           &5 &0.2543 &0.3980 &0.4656 &\textbf{0.9055}
                              &0.2476 &0.2980 &0.3245 &\textbf{0.6774}\\      
        \hline\hline
      \end{tabular}
    \end{small}
\end{table}

\subsection{Synthetic Data}\label{ssec:syn}
We generate two synthetic data sets with various clusters using sine-like impact functions and piecewise constant impact functions respectively. 
In each data set, the number of clusters is set from $2$ to $5$. 
Each cluster contains $400$ event sequences, and each event sequence contains $50$ ($=M_n$) events and $5$ ($=C$) event types. 
The elements of exogenous base intensity are sampled uniformly from $[0,1]$. 
Each sine-like impact function in the $k$-th cluster is formulated as $\phi_{cc'}^{k}=b_{cc'}^k(1-\cos(\omega_{cc'}^{k}(t-s_{cc'}^{k})))$, where $\{b_{cc'}^k, \omega_{cc'}^k, s_{cc'}^k\}$ are sampled randomly from $[\frac{\pi}{5},\frac{2\pi}{5}]$. 
Each piecewise constant impact function is the truncation of the corresponding sine-like impact function, i.e., $2b_{cc'}^k\times \mbox{round}(\phi_{cc'}^k/(2b_{cc'}^k))$.

\begin{figure}[t]
\centering
\subfigure[Sine-like impact function]{
\includegraphics[width=0.44\columnwidth]{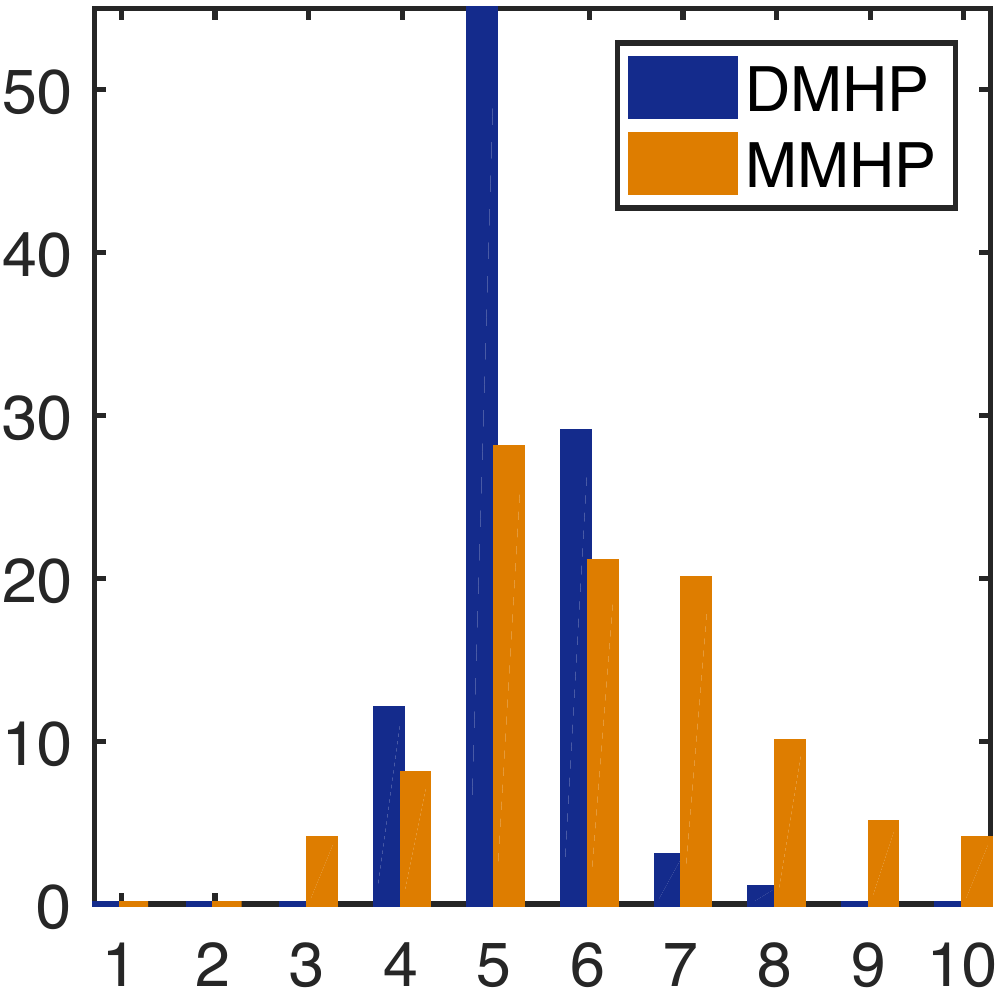}\label{fig:syn1}
}~~
\subfigure[Piecewise constant impact function]{
\includegraphics[width=0.44\columnwidth]{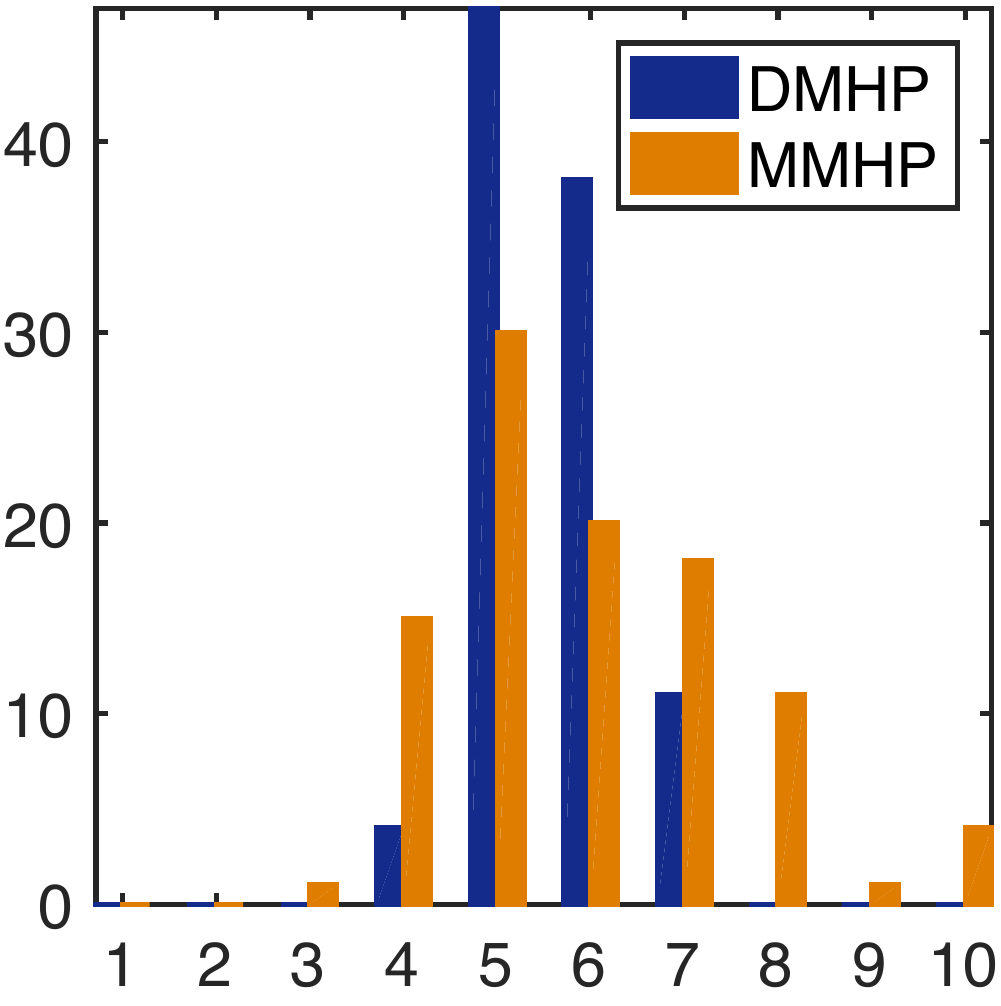}\label{fig:syn2}
}
\vspace{-10pt}
\caption{
The histograms of the number of clusters obtained via various methods on the two synthetic data sets.
}\label{fig:hist}\vspace{-3pt}
\end{figure}

Table~\ref{tab1} shows the clustering purity of various methods on the  synthetic data. 
Compared with the three competitors, our DMHP obtains much better clustering purity consistently. 
The VAR simply treats asynchronous event sequences as time series, which loses the information like the order of events and the time delay of adjacent events. 
Both the LS and the MMHP learn Hawkes process for each individual sequence, which might suffer to over-fitting problem in the case having few events per sequence. 
These competitors decompose sequence clustering into two phases: learning feature and applying DPGMM, which is very sensitive to the quality of feature. 
The potential problems above lead to unsatisfying clustering results. 
Our DMHP method, however, is model-based, which learns clustering result directly and reduces the number of unknown variables greatly. 
As a result, our method avoids the problems of these three competitors and obtains superior clustering results. 
Additionally, the learning results of the synthetic data with piecewise constant impact functions prove that our DMHP method is relatively robust to the problem of model misspecification --- although our Gaussian basis cannot fit piecewise constant impact functions well, our method still outperforms other methods greatly. 
Fig.~\ref{fig:hist} shows the histograms of the number of clusters obtained via various methods on our two synthetic data sets ($K=5$). 
We can find that the distributions obtained by our method are more concentrated to the real number of clusters. 

\subsection{Real-world Data}\label{ssec:real}
We test our clustering method on two real-world data sets. 
The first is the ICU patient flow data used in~\cite{xu2016icu}, which is extracted from the MIMIC II data set~\cite{saeed2002mimic}. 
This data set contains the transition processes of $30,308$ patients among different kinds of care units. 
The patients can be clustered according to their transition processes. 
The second is the IPTV data set in~\cite{luo2014you,luo2015multi}, which contains $7,100$ IPTV users' viewing records collected via Shanghai Telecomm Inc. 
The TV programs are categorized into $16$ classes and the viewing behaviors more than $20$ minutes are recorded. 
Similarly, the users can be clustered according to their viewing records. 
The event sequences in these two data have strong but structural triggering patterns, which can be modeled via different Hawkes processes.

\begin{table}[t]
  \centering
  \caption{Clustering Consistency on Real-world Data.\label{tab3}}\vspace{-5pt}
    \begin{small}
      \begin{tabular}{
        @{\hspace{3pt}}c@{\hspace{3pt}}|
        @{\hspace{3pt}}c@{\hspace{3pt}}
        @{\hspace{3pt}}c@{\hspace{3pt}}
        @{\hspace{3pt}}c@{\hspace{3pt}}
        @{\hspace{3pt}}c@{\hspace{3pt}}
        }
        \hline\hline
        Method &VAR+DPGMM &LS+DPGMM &MMHP+DPGMM &DMHP\\
        \hline
        ICU Patient &0.0901  &0.1390  &0.3313  &\textbf{0.3778}\\
        IPTV User   &0.0443  &0.0389  &0.1382  &\textbf{0.2004}\\      
        \hline\hline
      \end{tabular}
    \end{small}
\end{table}

\begin{figure}[t]
\centering
\begin{minipage}[b]{0.3\columnwidth}
\subfigure[Histogram of $K$]{
\includegraphics[width=0.9\columnwidth]{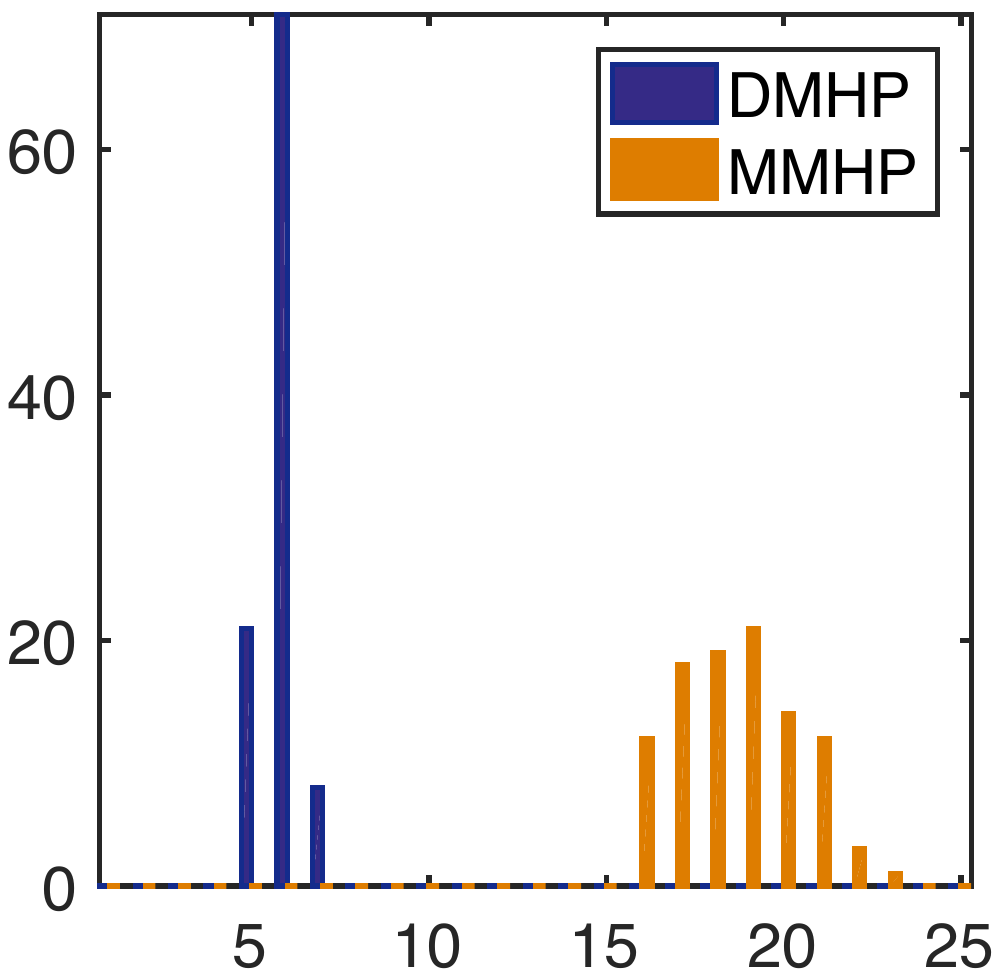}\label{fig:icu}
}~~
\end{minipage}
\begin{minipage}[b]{0.6\columnwidth}
\subfigure[DMHP]{
\includegraphics[width=1\columnwidth]{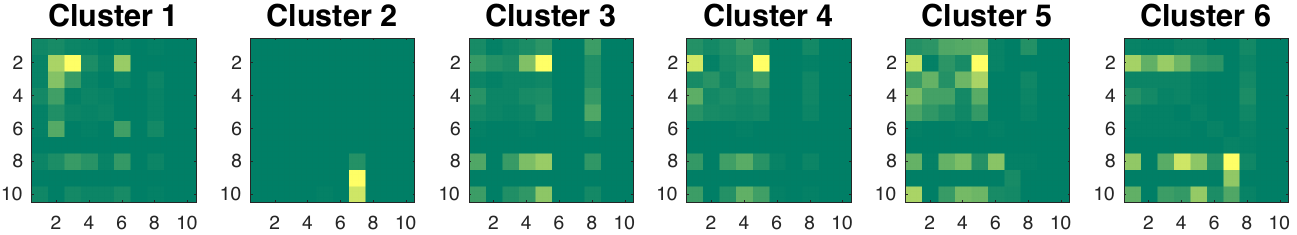}\label{fig:ICU1}
}\vspace{-5pt}\\
\subfigure[MMHP+DPGMM]{
\includegraphics[width=1\columnwidth]{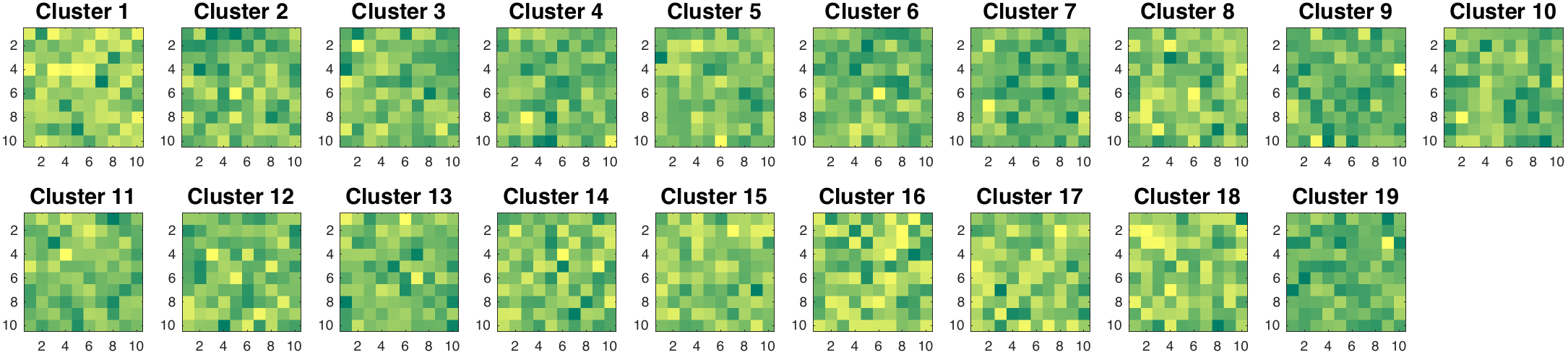}\label{fig:ICU2}
}
\end{minipage}
\vspace{-10pt}
\caption{
Comparisons on the ICU patient flow data.
}\label{fig:ICUL}\vspace{-3pt}
\end{figure}

\begin{figure}[t]
\centering
\begin{minipage}[b]{0.34\columnwidth}
\subfigure[Histogram of $K$]{
\includegraphics[width=1\columnwidth]{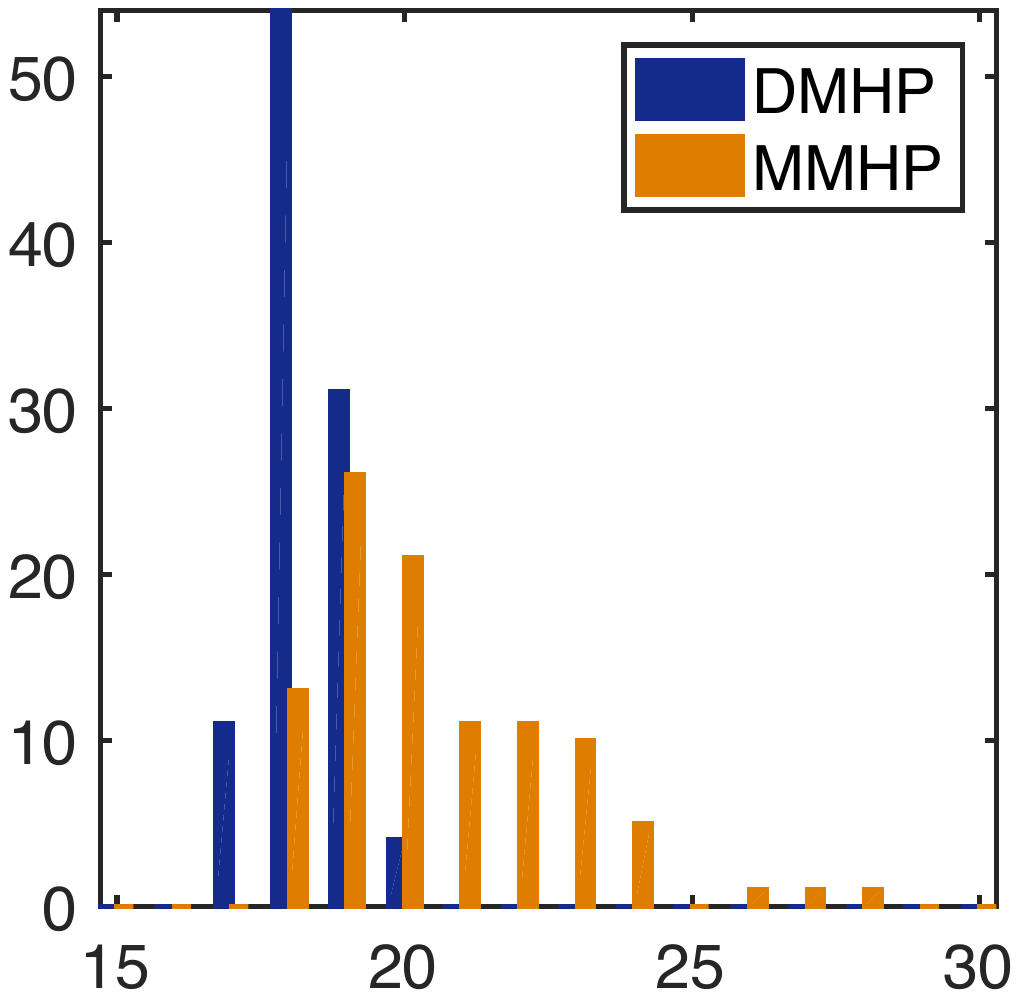}\label{fig:iptv}
}~~
\end{minipage}
\begin{minipage}[b]{0.62\columnwidth}
\subfigure[DMHP]{
\includegraphics[width=1\columnwidth]{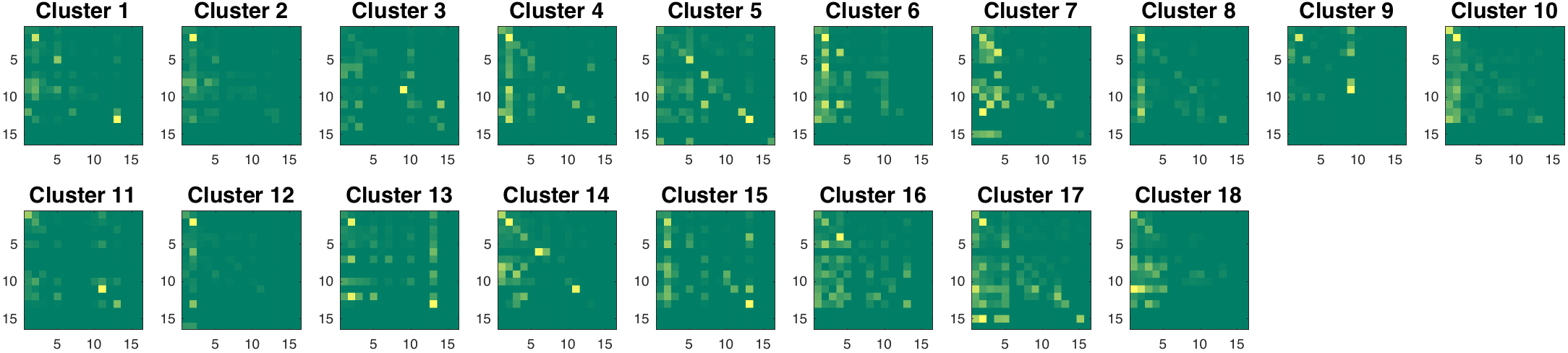}\label{fig:IPTV1}
}\vspace{-5pt}\\
\subfigure[MMHP+DPGMM]{
\includegraphics[width=1\columnwidth]{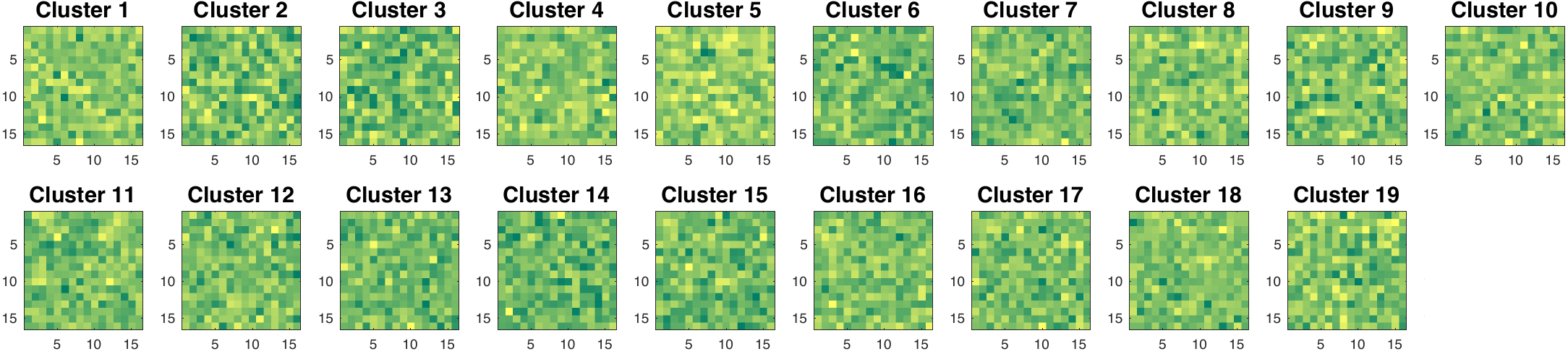}\label{fig:IPTV2}
}
\end{minipage}
\vspace{-10pt}
\caption{
Comparisons on the IPTV user data.
}\label{fig:IPTVL}\vspace{-3pt}
\end{figure}

Table~\ref{tab3} shows the performance of various clustering methods on the clustering consistency. 
We can find that our method outperforms other methods obviously, which means that the clustering result obtained via our method is more stable and consistent than other methods' results. 
In Fig.~\ref{fig:ICUL} we visualize the comparison for our method and its main competitor MMHP+DPGMM on the ICU patient flow data. 
Fig.~\ref{fig:icu} shows the histograms of the number of clusters for the two methods. 
We can find that MMHP+DPGMM method tends to over-segment data into too many clusters. 
Our DMHP method, however, can find more compact clustering structure. 
The distribution of the number of clusters concentrates to $6$ and $19$ for the two data sets, respectively. 
In our opinion, this phenomenon reflects the drawback of the feature-based method --- the clustering performance is highly dependent on the quality of feature while the clustering structure is not considered sufficiently in the phase of extracting feature. 
Taking learned infectivity matrices as representations of clusters, we compare our DMHP method with MMHP+DPGMM in Figs.~\ref{fig:ICU1} and~\ref{fig:ICU2}. 
The infectivity matrices obtained by our DMHP are sparse and with distinguishable structure, while those obtained by MMHP+DPGMM are chaotic --- although MMHP also applies sparse regularizer to each event sequence' infectivity matrix, it cannot guarantee the average of the infectivity matrices in a cluster is still sparse. 
Same phenomena can also be observed in the experiments on the IPTV data. 
The clustering results of IPTV data are shown in Fig.~\ref{fig:IPTVL}. 
Compared with the results obtained via MMHP+DPGMM, the histogram of the number of clusters obtained via our DMHP method is more concentrated and the infectivity matrices of clusters are more structural and explainable.

\section{Conclusion and Future Work}\label{sec:con}
In this paper, we propose and discuss a Dirichlet mixture model of Hawkes processes and achieve a model-based solution to event sequence clustering. 
We prove the identifiability of our model and analyze the convergence, sample complexity and computational complexity of our learning algorithm.
In the aspect of methodology, we plan to study other potential priors, e.g., the prior based on determinantial point processes (DPP) in~\cite{xu2016bayesian}, to improve the estimation of the number of clusters, and further accelerate our learning algorithm via optimizing inner iteration allocation strategy in near future. 
Additionally, our model can be extended to Dirichlet process mixture model when $K\rightarrow \infty$. 
In that case, we plan to apply Bayesian nonparametrics to develop new learning algorithms.

\section{Acknowledgment}
This work is supported in part by NSF IIS-1639792, IIS-1717916, and CMMI-1745382. 

\section{Supplementary File}
\subsection{The Proof of Local Identifiability}
Before proving the local identifiability of our DMHP model, we first introduce some key concepts. 
A temporal point process is a random process whose realization consists of a list of discrete events in time $\{ t_i \}$ with $t_i \in [0,T]$. 
Here $[0,T]$ is the time interval of the process.
It can be equivalently represented as a counting process,
$N=\{N(t)|t\in [0,T]\}$, where $N(t)$ records the number of events before time $t$.
A multi-dimensional point process with $C$ types of event is represented by $C$ counting processes $\{N_c\}_{c=1}^{C}$ on a probability space $(\Omega, \mathfrak{F}, \mathbb{P})$. 
$N_c=\{N_c(t)| t\in [0,T]\}$, where $N_c(t)$ is the number of type-$c$ events occurring at or before time $t$. 
$\Omega=[0,T]\times \mathcal{C}$ is the sample space. 
$\mathcal{C}=\{1,...,C\}$ is the set of event types. 
$\mathfrak{F}=(\mathfrak{F}(t))_{t\in\mathbb{R}}$ is the filtration representing the set of events sequence the process can realize until time $t$. 
$\mathbb{P}$ is the probability measure.

Hawkes process is a kind of temporal point processes having self-and mutually-triggering patterns. 
The triggering of historical events on current ones in a Hawkes process can be modeled as branch processes~\cite{simma2010modeling,farajtabar2014shaping}. 
As a result, Hawkes Process can be represented as a superposition of many non-homogeneous Poisson process. 
Due to the superposition theorem of Poisson processes, the superposition of the individual processes is equivalent to the point process with summation of their intensity function. 
Given this we can break the counting process associated to each addition to the intensity function (or associated to each event): $N(t) = \sum_{i=0}^n N^i(t)$, where $N^0(t)$ is the counting process associated to the baseline intensity $\mu(t)$ and $N^i(t)$ is the non-homgenous Poisson process for the $i$-th branch.
Similarly, we can write the intensity function of Hawkes process as $\lambda(t) = \sum_{i=0}^n \lambda^i(t)$, 
where $\lambda^i(t)$ is the intensity of the $i$-th branch.

\begin{defn}\label{def1}
\emph{Two parameter points $\bm{\Theta}^1$ and $\bm{\Theta}^2$ are said to be observationally equivalent if $p(\bm{s};\bm{\Theta}^1)=p(\bm{s};\bm{\Theta}^2)$ for all samples $\bm{s}$'s in sample space. }
\end{defn}

\begin{defn}\label{def2}
\emph{A parameter point $\bm{\Theta}^0$ is said to be locally identifiable if there exists an open neighborhood of $\bm{\Theta}^0$ containing no other $\bm{\Theta}$ in the parameter space which is observationally equivalent.}
\end{defn}

\begin{defn}\label{def3}
\emph{Let $\bm{I}(\bm{\Theta})$ be a matrix whose elements are continuous functions of $\bm{\Theta}$ everywhere in the parameter space. The point $\bm{\Theta}^0$ is said to be a regular point of the matrix if there exists an open neighborhood of $\bm{\Theta}^0$ in which $\bm{I}(\bm{\Theta})$ has constant rank.}
\end{defn}

The information matrix $\bm{I}(\bm{\Theta})$ is defined as
\begin{eqnarray*}
\begin{aligned}
\bm{I}(\bm{\Theta})=\mathbb{E}_{\bm{s}}\left[\frac{\partial \log p(\bm{s};\bm{\Theta})}{\partial \bm{\Theta}}\frac{\partial \log p(\bm{s};\bm{\Theta})}{\partial \bm{\Theta}^{\top}}\right]
=\mathbb{E}_{\bm{s}}\left[\frac{1}{p^2(\bm{s};\bm{\Theta})}\frac{\partial p(\bm{s};\bm{\Theta})}{\partial \bm{\Theta}}\frac{\partial p(\bm{s};\bm{\Theta})}{\partial \bm{\Theta}^{\top}}\right],
\end{aligned}
\end{eqnarray*}

The local identifiability of our DMHP model is based on the following two theorems. 
\begin{thm}\label{the2}
\cite{meijer2008simple} The information matrix $\bm{I}(\bm{\Theta})$ is positive definite if and only if there does not exist a nonzero vector of constants $\bm{w}$ such that $\bm{w}^{\top}\frac{\partial p(\bm{s};\bm{\Theta})}{\partial \bm{\Theta}}=0$ for all samples $\bm{s}$'s in sample space.
\end{thm}

\begin{thm}\label{the3}
\cite{rothenberg1971identification} Let $\bm{\Theta}^{0}$ be a regular point of the information matrix $\bm{I}(\bm{\Theta})$. 
Then $\bm{\Theta}^0$ is locally identifiable if and only if $\bm{I}(\bm{\Theta}^0)$ is nonsingular. 
\end{thm}

To our DMHP model, the log-likelihood function is composed with differentiable functions of $\bm{\Theta}$. 
Therefore, the elements of information matrix $\bm{I}(\bm{\Theta})$ are continuous functions w.r.t. $\bm{\Theta}$ in the parameter space. 
According to Theorems~\ref{the2} and~\ref{the3}, our Theorem holds if and only if to each vector $\frac{\partial p(\bm{s};\bm{\Theta})}{\partial \bm{\Theta}}$ w.r.t. a point $\bm{\Theta}$, there does not exist a nonzero vector of constants $\bm{w}$ such that $\bm{w}^{\top}\frac{\partial p(\bm{s};\bm{\Theta})}{\partial \bm{\Theta}}=0$ for all event sequences $\bm{s}\in\mathfrak{F}$. 

Assume that there exists a nonzero $\bm{w}$ such that $\bm{w}^{\top}\frac{\partial p(\bm{s};\bm{\Theta})}{\partial \bm{\Theta}}=0$ for all $\bm{s}\in\mathfrak{F}$. 
We have the following \textbf{counter-evidence:} Considering the simplest case --- the mixture of two Poisson processes (or equivalently, two 1-dimensional Hawkes processes whose impact functions $\phi(t)\equiv 0$), we can write its likelihood given a sequence with $N$ events in $[0,T]$ as
\begin{eqnarray*}
\begin{aligned}
p(\bm{s}_{N};\bm{\Theta})&=\pi\lambda_1^N\exp(-T\lambda_1)+(1-\pi)\lambda_2^N\exp(-T\lambda_2)
=\Lambda_1+\Lambda_2,
\end{aligned}
\end{eqnarray*}
where $\bm{\Theta}=[\pi,\lambda_1,\lambda_2]^{\top}$, $\lambda_1\neq\lambda_2$. 
According to our assumption, we have
\begin{eqnarray*}
\begin{aligned}
\bm{w}^{\top}\frac{\partial p(\bm{s}_N;\bm{\Theta})}{\partial \bm{\Theta}}
=\bm{w}^{\top}
\begin{bmatrix}
\frac{\Lambda_1}{\pi}-\frac{\Lambda_2}{1-\pi}\\
(\frac{N}{\lambda_1}-T)\Lambda_1\\
(\frac{N}{\lambda_2}-T)\Lambda_2
\end{bmatrix}
=0,
\end{aligned}
\end{eqnarray*}
Denote the time stamp of the last event as $t_N$, we can generate new event sequences $\{\bm{s}_{N+n}\}_{n=1}^{\infty}$ via adding $n$ events in $(t_N, T]$, and
\begin{eqnarray*}
\begin{aligned}
\bm{w}^{\top}\frac{\partial p(\bm{s}_{N+n};\bm{\Theta})}{\partial \bm{\Theta}}
=\bm{w}^{\top}
\begin{bmatrix}
\lambda_1^{n}\frac{\Lambda_1}{\pi}-\lambda_2^{n}\frac{\Lambda_2}{1-\pi}\\
((N+n)-T\lambda_1)\lambda_1^{n-1}\Lambda_1\\
((N+n)-T\lambda_2)\lambda_2^{n-1}\Lambda_2
\end{bmatrix}.
\end{aligned}
\end{eqnarray*}
$\bm{w}^{\top}\frac{\partial p(\bm{s}_{N+n};\bm{\Theta})}{\partial \bm{\Theta}}=0$ for $n=0,...,\infty$ requires $\bm{w}\equiv \bm{0}$ or all $\frac{\partial p(\bm{s}_{N+n};\bm{\Theta})}{\partial \bm{\Theta}}$ are coplanar. 
However, according to the formulation above, 
for arbitrary three different $n_1, n_2, n_3\in\{0,...,\infty\}$, $\sum_{i=1}^{3}\alpha_i\frac{\partial p(\bm{s}_{N+n_i};\bm{\Theta})}{\partial \bm{\Theta}}=\bm{0}$ holds if and only if $\alpha_1=\alpha_2=\alpha_3=0$.\footnote{The derivation is simple. Interested reader can try the case with $n_1=0$, $n_2=1$, $n_3=3$} 
Therefore, $\bm{w}\equiv\bm{0}$, which violates the assumption above.

Such a counter-evidence can also be found in more general case, i.e., mixtures of multiple multi-dimensional Hawkes processes because Hawkes process is a superposition of many non-homogeneous Poisson process. 
As a result, according to Theorems~\ref{the2} and~\ref{the3}, each point $\bm{\Theta}$ in the parameter space is regular point of $\bm{I}(\bm{\Theta})$ and the $\bm{I}(\bm{\Theta})$ is nonsingular, and thus, our DMHP model is locally identifiable.

\subsection{The Selection of Basis Functions}
In our work, we apply Gaussian basis functions to our model. 
We use the basis selection method in~\cite{xu2016learning} to decide the bandwidth and the number of basis functions. 
In particular, we focus on the impact functions having Fourier transformation. 
The representation of impact function, i.e., $\phi_{cc'}(t)=\sum_{d=1}^{D} a_{cc'}g_d(t)$, can be explained as a sampling process, where $\{a_{cc'}^{d}\}_{d=1}^{D}$ can be viewed as the discretized samples of $\phi_{cc'}(t)$ in $[0,T]$ and each $g_d(t)=\kappa_{\omega}(t,t_d)$ is sampling function with cut-off frequence $\omega$ and center $t_d$. 
Given training sequences $\bm{S}=\{\bm{s}_n=\{(t_i, c_i)\}_{i=1}^{M_n}\}_{n=1}^{N}$, we can estimate $\lambda(t)$ empirically via a Gaussian-based kernel density estimator:  
\begin{eqnarray}\label{intensityAll}
\begin{aligned}
\lambda(t)=\sideset{}{_{n=1}^{N}}\sum\sideset{}{_{i=1}^{M_n}}\sum G_h(t-t_i).
\end{aligned}
\end{eqnarray}
Here $G_h(t-t_i)=\exp(-\frac{(t-t_i)^2}{2h^2})$ is a Gaussian kernel with the bandwidth $h$. 
Instead of computing~(\ref{intensityAll}), we directly apply Silverman's rule of thumb~\cite{silverman1986density} to set optimal $h=(\frac{4\hat{\sigma}^5}{3\sum_n M_n})^{0.2}$, where $\hat{\sigma}$ is the standard deviation of time stamps $\{t_i\}$. 
Applying Fourier transform, we compute an upper bound for the spectral of $\lambda(t)$ as
\begin{eqnarray}
\begin{aligned}
|\hat{\lambda}(\omega)| &= \left| \int_{-\infty}^{\infty}\lambda(t)e^{-j\omega t}dt\right|=\left| \sideset{}{_{n=1}^{N}}\sum\sideset{}{_{i=1}^{M_n}}\sum\int_{-\infty}^{\infty}e^{-\frac{(t-t_i)^2}{2h^2}}e^{-j\omega t}dt \right|\\
&\leq \sideset{}{_{n=1}^{N}}\sum\sideset{}{_{i=1}^{M_n}}\sum\left| \int_{-\infty}^{\infty}e^{-\frac{(t-t_i)^2}{2h^2}}e^{-j\omega t}dt \right|=\sideset{}{_{n=1}^{N}}\sum\sideset{}{_{i=1}^{M_n}}\sum\left| e^{-j\omega t_i}e^{-\frac{\omega^2 h^2}{2}}\sqrt{2\pi h^2} \right|\\
&\leq \sideset{}{_{n=1}^{N}}\sum\sideset{}{_{i=1}^{M_n}}\sum\left| e^{-j\omega t_i }\right|\left|e^{-\frac{\omega^2 h^2}{2}}\sqrt{2\pi h^2} \right|=\left(\sideset{}{_{n=1}^{N}}\sum M_n\sqrt{2\pi h^2}\right)e^{-\frac{\omega^2 h^2}{2}}.
\end{aligned}
\end{eqnarray}
Then, we can compute the upper bound of the absolute sum of the spectral higher than a certain threshold $\omega_0$ as
\begin{eqnarray*}
\begin{aligned}
\int_{\omega_0}^{\infty}|\hat{\lambda}(\omega)|d\omega
\leq &\left(\sideset{}{_{n=1}^{N}}\sum M_n\sqrt{2\pi h^2}\right)\int_{\omega_0}^{\infty}e^{-\frac{\omega^2 h^2}{2}}d\omega
=\pi\left(\sideset{}{_{n=1}^{N}}\sum M_n\right)\left(1-\frac{1}{\sqrt{2}}\mbox{erf}(\omega_0 h)\right),
\end{aligned}
\end{eqnarray*}
where $\mbox{erf}(x)=\frac{1}{\sqrt{\pi}}\int_{-x}^{x}e^{-t^2}dt$. 

Therefore, give a bound of residual $\epsilon$, we can find an $\omega_0$ guaranteeing $\int_{\omega_0}^{\infty}|\hat{\lambda}(\omega)|d\omega\leq \epsilon$, or $\mbox{erf}(\omega_0 h)\geq \sqrt{2}-\frac{\sqrt{2}\epsilon}{\pi\sum_{n=1}^{N}M_n}$. 
The proposed basis functions $\{g_d(t)\}_{d=1}^{D}$ are selected --- each $g_d(t)$ is a Gaussian function with cut-off frequency $\omega_0$ and center $\frac{(d-1)T}{D}$, where $D=\lceil\frac{T\omega_0}{\pi}\rceil$. 
In summary, we propose Algorithm~\ref{algBasis} to select basis functions. 
\begin{algorithm}
   \caption{Selecting basis functions}
   \label{algBasis}
\begin{algorithmic}[1]
   \STATE \textbf{Input:} $\bm{S}=\{\bm{s}_n\}_{n=1}^{N}$, residual's upper bound $\epsilon$.
   \STATE \textbf{Output:} Basis functions $\{g_{d}(t)\}_{d=1}^{D}$.
   \STATE Compute $\left(\sum_{n=1}^{N}M_n\sqrt{2\pi h^2}\right)e^{-\frac{\omega^2 h^2}{2}}$ to bound $|\hat{\lambda}(\omega)|$. 
   \STATE Find the smallest $\omega_0$ satisfying $\int_{\omega_0}^{\infty}|\hat{\lambda}(\omega)| d\omega\leq \epsilon$. 
   \STATE The Gaussian basis functions $\{g_{d}(t)\}_{d=1}^{D}$ are with cut-off frequency $\omega_0$ and centers $\{\frac{(d-1)T}{D}\}_{d=1}^D$, where $D=\lceil\frac{T\omega_0}{\pi}\rceil$.
\end{algorithmic}
\end{algorithm}

\subsection{Nested EM Framework}
We consider a variational distribution having the following factorization:
\begin{eqnarray}
\begin{aligned}
q(\bm{Z},\bm{\pi},\bm{\mu},\bm{A})=q(\bm{Z})q(\bm{\pi},\bm{\mu},\bm{A})=q(\bm{Z})q(\bm{\pi})\sideset{}{_k}\prod q(\bm{\mu}^k)q(\bm{A}^k).
\end{aligned}
\end{eqnarray}
An nested EM algorithm can be used to optimize (\ref{surrogate}). 

\textbf{Update Responsibility (E-step).} In each {\it outer iteration}, the logarithm of the optimized factor $q^{*}(\bm{Z})$ is approximated as
\begin{eqnarray}
\begin{aligned}
&\log q^*(\bm{Z})\\
=&\mathbb{E}_{\bm{\pi},\bm{\mu},\bm{A}}[\log p(\bm{S},\bm{Z},\bm{\pi},\bm{\mu},\bm{A})]+\mathsf{C}\\
=&\mathbb{E}_{\bm{\pi}}[\log p(\bm{Z}|\bm{\pi})]+\mathbb{E}_{\bm{\mu},\bm{A}}[\log p(\bm{S}|\bm{Z},\bm{\mu},\bm{A})]+\mathsf{C}\\
=&\sideset{}{_{n,k}}\sum z_{nk}\left(\mathbb{E}[\log \pi^k]+\mathbb{E}[\log\mbox{HP}(\bm{s}_n|\bm{\mu}^k,\bm{A}^k)]\right)+\mathsf{C}\\
=&\sideset{}{_{n,k}}\sum z_{nk}\Bigl(\mathbb{E}[\log \pi^k]+\mathbb{E}[\sideset{}{_i}\sum\log\lambda_{c_i}^{k}(t_i)-\sideset{}{_c}\sum\int_0^{T_n}\lambda_c^k(s)ds]\Bigr)+\mathsf{C}\\
\approx &\sideset{}{_{n,k}}\sum z_{nk}\Bigr(\mathbb{E}[\log \pi^k]+\sideset{}{_i}\sum\Bigl(\log\mathbb{E}[\lambda_{c_i}^{k}(t_i)]-\frac{\text{Var}[\lambda_{c_i}^{k}(t_i)]}{2\mathbb{E}^2[\lambda_{c_i}^{k}(t_i)]}\Bigr)
-\sideset{}{_c}\sum\mathbb{E}[\int_0^{T_n}\lambda_c^k(s)ds]\Bigl)+\mathsf{C}\\
=&\sideset{}{_{n,k}}\sum z_{nk}\log \rho_{nk} + \mathsf{C}.
\end{aligned}
\end{eqnarray}
where $\mathsf{C}$ is a constant, and each term $\mathbb{E}[\log\lambda_{c}^{k}(t)]$ is approximated via its second-order Taylor expansion $\log\mathbb{E}[\lambda_{c}^{k}(t)]-\frac{\text{Var}[\lambda_{c}^{k}(t)]}{2\mathbb{E}^2[\lambda_{c}^{k}(t)]}$~\cite{teh2006collapsed}. 
Then, we have
\begin{eqnarray*}
\begin{aligned}
&\log\rho_{nk}\\
=&\mathbb{E}[\log \pi^k]+\sum_i\Bigl(\log(\mathbb{E}[\lambda_{c_i}^{k}(t_i)])-\frac{\text{Var}[\lambda_{c_i}^{k}(t_i)]}{2\mathbb{E}^2[\lambda_{c_i}^{k}(t_i)]}\Bigr)-\sum_c\mathbb{E}[\int_0^{T_n}\lambda_c^k(s)ds]\\
=&\mathbb{E}[\log \pi^k]+\sum_i\Bigl(\log(\mathbb{E}[\mu_{c_i}^{k}]
+\sum_{j<i,d}\mathbb{E}[a_{c_i c_j d}^{k}]g_d(\tau_{ij}))
-\frac{\text{Var}[\mu_{c_i}^{k}]+\sum_{j<i,d}\text{Var}[a_{c_ic_jd}^{k}]g_d^2(\tau_{ij})}{2(\mathbb{E}[\mu_{c_i}^{k}]
+\sum_{j<i,d}\mathbb{E}[a_{c_i c_j d}^{k}]g_d(\tau_{ij}))^2}\Bigr)\\
&-\sum_c(T_n\mathbb{E}[\mu_c^k]+\sum_{i,d}\mathbb{E}[a_{cc_i d}^{k}]G_d(T_n-t_i))\\
=&\mathbb{E}[\log \pi^k]+\sum_i\Bigl(\log(\sqrt{\frac{\pi}{2}}\beta_{c_i}^{k}
+\sum_{j<i,d}\sigma_{c_i c_j d}^{k}g_d(\tau_{ij}))-\frac{\frac{4-\pi}{2}(\beta_{c_i}^{k})^2+\sum_{j<i,d}(\sigma_{c_ic_jd}^{k}g_d(\tau_{ij}))^2}{2(\sqrt{\frac{\pi}{2}}\beta_{c_i}^{k}
+\sum_{j<i,d}\sigma_{c_i c_j d}^{k}g_d(\tau_{ij}))^2}\Bigr)\\
&-\sum_c (T_n\sqrt{\frac{\pi}{2}}\beta_c^k + \sum_{i,d}\sigma_{cc_i d}^{k}G_d(T_n-t_i)),
\end{aligned}
\end{eqnarray*}
where $G_d(t)=\int_0^t g_d(s)ds$ and $\tau_{ij}=t_i-t_j$. 
The second equation above is based on the prior that all of the parameters are independent to each other. 
The term $\mathbb{E}[\log \pi^k]=\psi(\alpha_k)-\psi(\sum_k\alpha_k)$, where $\psi(\cdot)$ is the digamma function.\footnote{Denote the gamma function as $\Gamma(t)=\int_{0}^{\infty}x^{t-1}e^{-x}dx$, the digamma function is defined as $\psi(t)=\frac{d}{dt}\ln\Gamma(t)$.} 
Then, the responsibility $r_{nk}$ is calculated as
\begin{eqnarray}
\begin{aligned}
r_{nk}=\mathbb{E}[z_{nk}]=\frac{\rho_{nk}}{\sum_j\rho_{nj}},~\mbox{and}~N_k=\sideset{}{_n}\sum r_{nk}.
\end{aligned}
\end{eqnarray}
It should be noted that here we increase $q^*(\bm{Z})$ via maximizing its upper bound in each iteration because the difference between $q^*(\bm{Z})$ and its upper bound is bounded tightly. 
In particular, $q^*(\bm{Z})$ in~(\ref{qZ}) involves $\mathbb{E}[\log\lambda_{c_i}^k(t_i)]$, which is approximated via Jensen's inequality as $\log\mathbb{E}[\lambda_{c_i}^k(t_i)]$. 
It actually is the first order Talyor expansion of $\mathbb{E}[\log\lambda_{c_i}^k(t_i)]$. 
The second order term is bounded well and the higher order terms can be ignored.
We prove the rationality of our relaxation in the appendix.

\textbf{Update Parameters (M-step).} The optimal factor $q^*(\bm{\pi},\bm{\mu},\bm{A})$ is 
\begin{eqnarray}
\begin{aligned}
&\log q^*(\bm{\pi},\bm{\mu},\bm{A})\\
=&\sum_k\log(p(\bm{\mu}^k)p(\bm{A}^k)) + \mathbb{E}_{\bm{Z}}[\log p(\bm{Z}|\bm{\pi})]
+\log p(\bm{\pi})+\sum_{n,k} r_{nk}\log \mbox{HP}(\bm{s}_n|\bm{\mu}^k,\bm{A}^k)+\mathsf{C}.
\end{aligned}
\end{eqnarray}
We can estimate the parameters of Hawkes processes via:
\begin{eqnarray*}
\begin{aligned}
\max_{\bm{\mu},\bm{A}}~\log(p(\bm{\mu})p(\bm{A}))+\sideset{}{_{n,k}}\sum r_{nk}\log \mbox{HP}(\bm{s}_n|\bm{\mu}^k,\bm{A}^k).
\end{aligned}
\end{eqnarray*}
Here, we need to use an iterative method to solve the above optimization problem. 
Specifically, we initialize $\bm{\mu}$ and $\bm{A}$ via the expectations of their distributions (used in E-step), i.e., $\bm{\mu}=\sqrt{\frac{\pi}{2}}\bm{B}$ and $\bm{A}=\bm{\Sigma}$. 
Applying the Jensen's inequality, we obtain the surrogate function of the objective function:
\begin{eqnarray*}\label{surrogate2}
\begin{aligned}
&\log(p(\bm{\mu})p(\bm{A}))+\sum_{n,k} r_{nk}\log \mbox{HP}(\bm{s}_n|\bm{\mu}^k,\bm{A}^k)\\
=&\sum_{c,k}\left[\log\mu_c^k-\frac{1}{2}(\frac{\mu_c^k}{\beta_c^k})^2\right]-
\sum_{c,c',d,k}\frac{a_{cc'd}^{k}}{\sigma_{cc'd}^{k}}
+\sum_{n,k} r_{nk}\left[\sum_i\log\lambda_{c_i}^{k}(t_i)-\sum_c\int_0^{T_n}\lambda_{c}^{k}(s)ds \right]\\
\geq &\sum_{c,k}\left[\log\mu_c^k-\frac{1}{2}(\frac{\mu_c^k}{\beta_c^k})^2\right]-
\sum_{c,c',d,k}\frac{a_{cc'd}^{k}}{\sigma_{cc'd}^{k}}
+\sum_{n,k} r_{nk}\biggl[\sum_i\biggl( p_{ii}^k\log\frac{\mu_{c_i}^k}{p_{ii}}
+\sum_{j<i,d} p_{ijd}^k\log\frac{a_{c_i c_j d}^{k}g_{d}(\tau_{ij})}{p_{ijd}}\biggr)\\
&-\sum_c T_n\mu_{c}^{k} -\sum_{c,i,d} a_{cc_i d}^{k}G_d(T_n-t_i) \biggr]=Q,
\end{aligned}
\end{eqnarray*}
where $p_{ii}^k=\frac{\mu_{c_i}^k}{\lambda_{c_i}^k(t_i)}$, and $p_{ijd}^k=\frac{a_{c_i c_j d}^{k}g_d(\tau_{ij})}{\lambda_{c_i}^k(t_i)}$.
Setting $\frac{\partial Q}{\partial \mu_c^k}=0$ and $\frac{\partial Q}{\partial a_{cc'd}^k}=0$, we have
\begin{eqnarray}\label{Amu}
\begin{aligned}
\hat{\mu}_c^k = \frac{-b+\sqrt{b^2-4ac}}{2a},\quad
\hat{a}_{cc'd}^{k}=\frac{\sum_n r_{nk}\sum_{i:c_i=c}\sum_{j:c_j=c'} p_{ijd}^k}{1/\sigma_{cc'd}^k +\sum_n r_{nk}\sum_{i:c_i=c'}G_d(T_n-t_i)}.
\end{aligned}
\end{eqnarray}
where $a=\frac{1}{(\beta_c^k)^{2}}$,~$b=\sum_n r_{nk}T_n$,~$c=-1-\sum_n r_{nk}\sum_{i:c_i=c} p_{ii}^k$. 
After repeating several such {\it inner iterations}, we can get optimal $\hat{\bm{\mu}}$, $\widehat{\bm{A}}$, and update distributions as
\begin{eqnarray}
\begin{aligned}
\bm{\Sigma}^k = \widehat{\bm{A}}^k,~
\bm{B}^k =\sqrt{2/\pi}\hat{\bm{\mu}}^k.
\end{aligned}
\end{eqnarray}
The distribution of clusters can be estimated via $\pi^k = \frac{N_k}{N}$.

\subsection{Update The Number of Clusters $K$ via MCMC} 
In the case of infinite mixture model, we can apply the Markov chain Monte Carlo (MCMC)~\cite{green1995reversible,zhang2004learning,xu2016bayesian} to update $K$ via merging or splitting clusters. 

\textbf{Chose move type.} We make a random choice to propose a combine or a split move. 
Let $q_m$ and $q_s = 1-q_m$ denote the probability of proposing a merge and a split move, respectively, for a current $K$. 
Following the work in~\cite{xu2016bayesian}, we use $q_m =0.5$ for $K\geq 2$, and $q_m =0$ for $K=1$. 

\textbf{Merge move.} We randomly select a pair $(k_1,k_2)$ of components to merge and form a new component $k$. 
The probability of choosing $(k_1,k_2)$ is $q_c(k_1, k_2) = \frac{1}{K(K-1)}$. 
For our model, we can apply the following deterministic transformation to get new merged parameters:
\begin{eqnarray}
\begin{aligned}
\pi^k = \pi^{k_1}+\pi^{k_2},\quad
\bm{A}^{k} = \frac{\pi^{k_1}}{\pi^{k}}\bm{A}^{k_1}+\frac{\pi^{k_2}}{\pi^{k}}\bm{A}^{k_2},\quad
\bm{\mu}^{k} = \frac{\pi^{k_1}}{\pi^{k}}\bm{\mu}^{k_1}+\frac{\pi^{k_2}}{\pi^{k}}\bm{\mu}^{k_2}.
\end{aligned}
\end{eqnarray}
Then $\bm{\Sigma}$ and $\bm{B}$ are updated accordingly. 

\textbf{Split move.} We randomly select a component $k$ to split into two new components $k_1$ and $k_2$. 
The probability of choosing component $k$ is $q_s(k) = \frac{1}{K}$. 
Different from the sampling method in previous work~\cite{green1995reversible,zhang2004learning,xu2016bayesian}, the splitting of parameters is an ill-posed problem with positive constraints. 
Here, we apply a simple heuristic transformation to get new splitting parameters:
\begin{eqnarray}
\begin{aligned}
&\pi^{k_1} = a\pi^k,~\pi^{k_2} = (1-a)\pi^k,~ a\sim Be(1,1),\\
&\bm{A}^{k_1} = \frac{1}{2a}\bm{A}^k,~\bm{A}^{k_2} = \frac{1}{2(1-a)}\bm{A}^k,\quad
\bm{\mu}^{k_1} = \frac{1}{2a}\bm{\mu}^k,~\bm{\mu}^{k_2} = \frac{1}{2(1-a)}\bm{\mu}^k.
\end{aligned}
\end{eqnarray}
Then $\bm{\Sigma}$ and $\bm{B}$ are updated accordingly. 

\textbf{Acceptance.} Given original parameters $\bm{\Theta}$ and the new $\bm{\Theta}'$, we accept a merge/split move with the probability $\min\{1, \mbox{likelihood ratio}\times\frac{p(\bm{\Theta}')}{p(\bm{\Theta})}\}$.

\bibliographystyle{ieee}
\bibliography{example_paper}

\end{document}